\documentclass{article}

\usepackage{tikz}

\usepackage{xcolor}
\usepackage{dsfont}
\usepackage{numprint}
\usepackage{subfig}
\usepackage{caption}
\usepackage[export]{adjustbox}
\usepackage{subcaption}
\usepackage{url}
\usepackage{setspace}
\usepackage{tabularx}

\usepackage{indentfirst}
\usepackage{amsthm,amsmath}
\usepackage[ruled,lined,linesnumbered]{algorithm2e}  
\usepackage{algorithmic}
\usepackage{booktabs}
\usepackage{nicematrix}
\usepackage{mathrsfs}
\usepackage{graphicx}
\usepackage{pgfplots}
\pgfplotsset{compat=1.17}

\usepackage{color}
\usepackage{verbatim}
\usepackage{textcomp}

\usepackage{bbold}

\usepackage[utf8]{inputenc}
\usepackage{array}
\usepackage{makecell}
\usepackage{wrapfig}

\usepackage{multirow}

\usepackage{soul}
\usepackage{balance}

\theoremstyle{definition}

\usepackage[english]{babel}
\usepackage{blindtext}
\usepackage[normalem]{ulem}

\usepackage{enumitem,kantlipsum}

\usepackage{romannum}

\usepackage{ulem}

\newcommand{\name}{$\mathtt{ItD}\ $}

\newcommand{\thetao}{\theta_{\mathrm{o}}}

\newenvironment{packeditemize}{
\begin{list}{$\bullet$}{
\setlength{\labelwidth}{8pt}
\setlength{\itemsep}{0pt}
\setlength{\leftmargin}{\labelwidth}
\addtolength{\leftmargin}{\labelsep}
\setlength{\parindent}{0pt}
\setlength{\listparindent}{\parindent}
\setlength{\parsep}{0pt}
\setlength{\topsep}{3pt}}}{\end{list}}

\usepackage[accepted]{icml2025}

\theoremstyle{plain}

\theoremstyle{definition}

\theoremstyle{remark}

\usepackage[textsize=tiny]{todonotes}

\begin{document}

\twocolumn[
% \icmltitle{Interpret-then-deactivate: Precise and Expandable Concept Erasing for T2I Diffusion Models with Sparse Autoencoder }
\icmltitle{Sparse Autoencoder as a Zero-Shot Classifier for Concept Erasing in Text-to-Image Diffusion Models}




\icmlsetsymbol{equal}{*}
% \icmlsetsymbol{ca}{$\dag$}
\icmlsetsymbol{intern}{$\dag$}

\begin{icmlauthorlist}
\icmlauthor{Zhihua Tian}{intern,zju}
\icmlauthor{Sirun Nan}{nus}
\icmlauthor{Ming Xu}{nus}
\icmlauthor{Shengfang Zhai}{pku}
\icmlauthor{Wenjie Qu}{nus}
\icmlauthor{Jian Liu}{zju}
% \icmlauthor{Kui Ren}{zju}
\icmlauthor{Ruoxi Jia}{equal,vt}
\icmlauthor{Jiaheng Zhang}{equal,nus}
\end{icmlauthorlist}

\icmlaffiliation{zju}{Zhejiang University}
\icmlaffiliation{nus}{National University of Singapore}
\icmlaffiliation{pku}{Peking University}
\icmlaffiliation{vt}{Virginia Tech}
\icmlcorrespondingauthor{Jian Liu}{liujian2411@zju.edu.cn}


\icmlkeywords{Machine Learning, ICML}

\vskip 0.3in
] 



\printAffiliationsAndNotice{\textsuperscript{*} Equal advising \textsuperscript{$\dag$} Work done while visiting NUS }
% \printAffiliationsAndNotice{$\dag$ Corresponding author.}
% %\icmlEqualContribution

\begin{abstract}
% Diffusion models (DM) have demonstrated remarkable performance in text-to-image generation, yet they pose ethic and practical concerns, such as generating unsafe, misleading content and copyright disputes. 

% Machine unlearning (MU) in diffusion models (DMs) aims to remove undesirable generative capabilities, such as the production of unsafe, misleading, or copyright-infringing content. However, due to the intricate entanglement of concepts within DM, unlearning a specific concept often results in the degradation of the model's ability to perform normal generation tasks. Existing approaches solve the problem by fine-tuning on a remaining dataset that contains additional text prompts. However, little has been done to understand what prompts could lead to better retainability.
% and we ask is it possible to elinimate the requirements of addtional prompts?

% In this paper, we make the first attempt to answer this question by proposing and investigating a new characteristic of machine unlearning for DM, namely \textit{concept entanglement}, which explains why erasing one concept would affect the generation of another concept. Build upon the character, we propose \todo{name}, a novel technique that implement pinpoint machine unlearning while does not degrade the capability of normal concept generation. Extensive evaluations
Text-to-image (T2I) diffusion models have achieved remarkable progress in generating high-quality images but also raise people's concerns about generating harmful or misleading content.
% (e.g., pornographic or copyrighted content). 
While extensive approaches have been proposed to erase unwanted concepts without requiring retraining from scratch, they inadvertently degrade performance on normal generation tasks. 
% Concept Erasing in text-to-image diffusion models aims to remove unwanted concepts absorbed during training without requiring retraining from scratch. 
% Concept Erasing is widely used in training text-to-image diffusion models to prevent harmful or misleading image generation (e.g., copyrighted or pornographic content). 
% While effective, it inadvertently degrades performance on normal generation tasks. 
% Existing approaches try to balance this by introducing a regularization term to minimize disruptions to normal concepts. However, due to the enormous scale of normal concepts, the regularization loss is usually under-trained, limiting its ability to preserve unseen concepts.
% Existing approaches to concept erasing typically rely on regularization techniques aimed at minimizing disruptions to normal concepts. However, the exponential growth of normal concepts 
% in recent models 
% often leads to under-trained regularization, which compromises their ability to preserve these concepts effectively. 
% Including additional concepts in the regularization term can partially address this issue but may inadvertently exacerbate degradation in other concepts. 
% \xm{It could be challenging to guarantee xx and xx.}
In this work, we propose \textbf{Interpret then Deactivate} ($\mathtt{ItD}$), a novel framework to enable precise concept removal in T2I diffusion models while preserving overall performance. 
\name first employs a sparse autoencoder (SAE) to interpret each concept as a combination of multiple features.
% and then identifies specific features associated with target concepts. 
By permanently deactivating the specific features associated with target concepts, we repurpose SAE as a zero-shot classifier that identifies whether the input prompt includes target concepts, allowing selective concept erasure in diffusion models. Moreover, we demonstrate that \name can be easily extended to erase multiple concepts without requiring further training.
% in the activation space, decomposing it into unique directions and shared directions. 
% By deactivating only the unique directions of target concepts during generation, ItD prevents their appearance while maintaining normal concepts unaffected. Furthermore, we 
Comprehensive experiments across celebrity identities, artistic styles, and explicit content demonstrate \name's effectiveness in eliminating targeted concepts without interfering with normal concept generation. Additionally, \name is also robust against adversarial prompts designed to circumvent content filters. Code is available at: \url{https://github.com/NANSirun/Interpret-then-deactivate}.

\end{abstract}

%-------------------------------------------------------------------------------

\section{Introduction}  
% Text-to-image (T2I) diffusion models have achieved remarkable success in generating high-quality images that 
% faithfully 
% reflect the input text descriptions
% ~\cite{dhariwal2021diffusion, saharia2022photorealistic, ruiz2023dreambooth, rombach2022high, chang2023muse}, \xm{and are widely used in xxx and xxx scenarios.}  
Text-to-image (T2I) diffusion models have achieved remarkable success in generating 
% high-quality 
images that faithful reflect the input text descriptions~\cite{dhariwal2021diffusion, saharia2022photorealistic, ruiz2023dreambooth, rombach2022high} while simultaneously raising 
% people's 
concerns about being used to generate images containing inappropriate content such as offensive, pornographic, copyrighted, or Not-Safe-For-Work (NSFW) content~\cite{schramowski2023safe,rando2022red,gandikota2023erasing}.
% , preventing their large-scale practical development. 
To mitigate the issue, concept erasing has been proposed to prevent T2I diffusion models from generating images relevant to the unwanted concepts without requiring retraining from scratch.
% that can be labeled as the to-be-erased concept. 
% For instance, erasing the concept of 'dog' would disable the model from generating images belonging to dog, even when prompted with the text 'dog' or other dog-related texts.  
% For instance, erasing the concept of ``nudity'' would disable the model from generating images semantically related to nudity, even when prompted with the word ``nudity'' or other related texts.  

A recent line of research proposes fine-tuning model parameters to remove the unwanted knowledge learned by diffusion models~\cite{gandikota2023erasing,kumari2023ablating,fan2023salun,huang2023receler,gandikota2024unified,orgad2023editing}. However, even only modifying the cross-attention (CA) layers within diffusion models would \textbf{inadvertently degrade the generation quality of normal concepts}~\cite{gandikota2023erasing,huang2023receler,zhang2024defensive,bui2024erasingadversarial}.
% In particular, these works demonstrate that fine-tuning the U-net module~\cite{ho2020denoising}, which works as a noise predictor within the diffusion model to predict noise added during the diffusion process, 
% can prevent the model from generating images semantically related to target concepts~\cite{gandikota2023erasing,ho2020denoising, jolicoeur2020adversarial}. 
% While fine-tuning effectively erases target concepts from the model~\cite{kumari2023ablating, zhang2024defensive, huang2023receler}, it 
For instance, erasing the concept of ``nudity'' could impair the model's ability to generate an image of a person. 
% as illustrated in Figure\todo{add figure}. 
To mitigate the issue, some approaches incorporate regularization techniques during fine-tuning to preserve the generation quality of normal concepts~\cite{kumari2023ablating,fan2023salun,ko2024boosting}. However, fine-tuning with a subset of concepts may introduce new biases into the model, leading to unpredictable performance degradation when generating other concepts~\cite{bui2024erasingadversarial}.
% Unfortunately, due to the enormous scale of normal concepts~\cite{ko2024boosting}, only a limited subset can be included in the regularization during training, leaving its effectiveness on unseen and unpredictable concepts uncertain.

As an alternative solution, some approaches integrate customized modules into the model to enable concept erasing without modifying the original model parameters\cite{lyu2024one, lu2024mace,conceptpinpoint2025}. 
% \citet{lyu2024one} and \citet{lu2024mace} leverage LoRA fine-tuning~\cite{hu2021lora} to modify the CA layers within the U-Net module. \citet{conceptpinpoint2025} integrated a gate module into each component (i.e., query, key, and value) within CA layers to selectively perform concept erasing. 
However, each module may also affect the generation of normal concepts due to its limited generalization capability. Moreover, erasing new concepts requires training additional modules, resulting in substantial computational overhead.

In this work, we aim to overcome the above limitation by introducing a novel framework, Interpret-then-Deactivate (\name), to enable  \textbf{precise} and \textbf{expandable} concept erasure in T2I diffusion models. \textbf{Precise} refers to the erasure only influences the generation of the target concept while \textbf{Expandable} means that the approach can be easily extended to erase multiple concepts without further training.

% while maintaining the generation performance of the remaining concepts. 
% does not degrade the performance of other concepts, 
% while 

\name employs sparse autoencoder (SAE)~\cite{olshausen1997sparse}, an unsupervised model, to learn the sparse features that constitute the semantic space of the text encoder. Within this space, we interpret each concept as a linear combination of sparse features, which may overlap with the feature sets between the target and normal concepts. 
We hypothesize that this overlap is a key factor causing unintended effects on the normal concepts during erasure.

% Building on this assumption, 
To this end, we propose to selectively erase the features unique to the target concept, enabling the precise erasure. 
% concept erasure. 
This can be achieved by first encoding the text embedding of the concept into the sparse feature space, deactivating the specific features, and then decoding it back into the embedding space.
To enable expandability in erasing multiple concepts, we can deactivate concept-specific features without requiring additional retraining. 
% To achieve the expandable ability when erasing multiple concepts, we can simply deactivate features specific to the concepts without requiring further retaining. 

In summary, we make the following contributions: (1) We propose \name, a novel framework to enable  \textbf{precise} and \textbf{expandable} concept erasure in T2I diffusion models. (2) To the best of our knowledge, we are the first to adopt SAE to concept erasing tasks in T2I diffusion models. (3) Extensive experiments across various datasets demonstrate that \name effectively erases target concepts while preserving the diversity of remaining concepts, outperforming baselines by large margins.

\section{Related Works}

\subsection{Sparse Autoencoder (SAE)}
\label{ssec:sae}
The internal of the neural network (NN) is hard to explain due to its polysemanticity nature, where neurons appear to activate in multiple, semantically distinct contexts. Recently, SAE has emerged as an effective tool for interpreting mechanisms of NN by breaking done the intermediate results into features interpretable to specific concepts~\cite{SparseAutoencodersInterpretable,kissane2024attentionlayersae}. Let $\mathbf{x}\in \mathbb{R}^{d_{in}}$ denote the input vector, the decoder and encoder of SAE can be formalized as:

\begin{equation}\nonumber
\footnotesize
\begin{aligned}
    \mathbf{z} & =\operatorname{ReLU}(W_{\text{enc}} \mathbf{x}+\mathbf{b}) \\
    \hat{\mathbf{x}} & =W_{\text{dec}} \mathbf{z} \\
    & =\sum_{i=0}^{d_{\text {nid }}-1} z_i \mathbf{f}_i
\end{aligned}
\end{equation}
where $W_{\text{enc}} \in \mathbb{R}^{d_{\text {hid}} \times d_{\text{in}}}$ and $W_{\text{dec}} \in \mathbb{R}^{d_{\text{in}} \times d_{\text{hid}}}$ are learned matrices of encoder and decoder.
$\mathbf{b} \in \mathbb{R}^{d_{\text{hid}}}$ is the learned bias. 
% Each column within $W_{\text{dec}}$ is treated as a learned feature. 
The loss to train the autoencoder is $\mathcal{L}(\mathbf{x})=\|\mathbf{x}-\hat{\mathbf{x}}\|_2^2+\alpha\mathcal{L}_{aux}$, where $\mathcal{L}_{aux}$ is the loss to control the sparsity of the reconstruction~\cite{SparseAutoencodersInterpretable} or to prevent dead features that do not been fired on a large number of training samples~\cite{gao2024scalingevaluatingsparseautoencoders}, scaled with coefficient $\alpha$. 

In our work, $\mathcal{L}_{aux}$ is the reconstruction error using only the largest $K_{\text{aux}}$ feature activations following~\cite{gao2024scalingevaluatingsparseautoencoders}. We adopt SAE to identify and deactivate features specific to the target concepts to disable the diffusion model to generate related images.

% Sparse autoencoder is a kind of neural network that aims to learn interpretable features by breaking down the intermediate results of models. 

% \subsection{Safe T2I image generation} 
\subsection{Concept Erasing in T2I Diffusion Model}
Diffusion models can be used to generate inappropriate content~\cite{zhang2025generate,schramowski2023safe}.
% one approach is data censoring 
% various approaches have been studied, such as dataset censoring
% ~\cite{sd2}, post-generation filtering~\cite{safetychecker}, or employing safety generation guidance~\cite{schramowski2023safe}. 
% To mitigate the concerns that T2I models are being used to generate unsafe contents, 
To address this issue, several approaches have been proposed, such as dataset censoring~\cite{sd2}, post-generation filtering~\cite{safetychecker}, and safety-guided generation~\cite{schramowski2023safe}. However, these methods either demand significant computational resources~\cite{sd2}, introduce new biases, or remain vulnerable to adversarial prompts~\cite{yang2024sneakyprompt, rando2022red}.
% or can be easily circumvented by adversarial inputs~\cite{rando2022red,yang2024sneakyprompt}. 

% \subsection{Concept erasing in T2I diffusion model}
To address these limitations, fine-tuning-based approaches have been extensively explored to erase target concepts. 
FMN~\cite{zhang2024forget} efficiently erases target concepts by re-steering cross-attention (CA) layers. 
UCE~\cite{gandikota2024unified} and TIME~\cite{orgad2023editing} modify the projection layer within CA layers with a closed-form solution. 
ESD~\cite{gandikota2023erasing} reduces the probability of generating images that are labeled as target concepts, and AC~\cite{kumari2023ablating} aligns the distribution of target concepts with surrogate concepts for concept erasure. To improve the effectiveness of erasing, SalUn~\cite{fan2023salun} and Scissorhands~\cite{wu2025scissorhands}  identify the most sensitive neurons related to target concepts and update only those neurons. 

\subsection{Forgetting on Remaining Concepts} 
While the above approaches demonstrate good performance in erasing target concepts, they inadvertently degrade the generation of remaining concepts~\cite{zhangunlearncanvas}. To alleviate the issue, many approaches~\cite{huang2023receler,heng2024selective, ko2024boosting,zhang2024defensive} utilize a regularization loss on remaining concepts to preserve their generation capability.
EAP~\cite{bui2024erasingadversarial} further investigate the impact of selecting different concepts for regularization, and OTE~\cite{bui2024adaptive}, which employs a similar training objective to AC~\cite{kumari2023ablating}, adaptively selecting the optimal surrogate concept during erasing. However, the effectiveness of regularization on unseen concepts remains unclear due to the enormous scale of normal concepts.

As an alternative solution, some approaches incorporate customized modules into the intermediate layers of the diffusion model without modifying original model parameters. we summarize them inference-based approaches.
% SPM~\cite{lyu2024one} and MACE~\cite{lu2024mace} incorporate LoRA modules~\cite{hu2021lora} into the intermediate layers of the diffusion model, enabling concept erasure without significantly altering the original model parameters.
SPM~\cite{lyu2024one} applies one-dimensional LoRA~\cite{hu2021lora} to the intermediate layers of diffusion models and proposes an anchoring loss for distant concepts. MACE~\cite{lu2024mace} propose using LoRA for erasing each target concept and introduced a loss integrating LoRAs from multiple target concepts, enabling the massive concept erasing while mitigating forgetting of remaining concepts.
CPE~\cite{conceptpinpoint2025} introduces a residual attention gate, a module inserted into each CA layer of the diffusion model to control whether erasure should be applied to a given concept. However, erasing multiple concepts requires training separate modules for each target concept, resulting in significant computational overhead.

% \subsection{Sparse Autoencoders}

% % gate-based approaches propose to selectively erase target concepts while maintaining the 
% % consistency 
% generation ability of remaining concepts. Lyu et al. showed the SPM~\cite{lyu2024one} that utilizes a one-dimensional LoRA plugged into pre-trained 

% \xm{DMs} to selectively erase specific learned concepts from the model's representations. 
% % learn the erasure for a specific concept. 
% Lu et al. proposed the MACE approach~\cite{lu2024mace} to support the mass concept erasure in diffusion models by leveraging cross-attention refinement along with LoRA fine-tuning.  
% % introduces a loss integrating LoRAs from multiple target concepts, enabling the massive concept erasing while mitigating forgetting of remaining concepts. 
% Later, CPE~\cite{conceptpinpoint} was proposed to 
% % theoretically 
% study the limitations of merely updating the cross-attention layers and proposes a nonlinear additive module to act as a gate that can recognize a target concept and instruct it to align with surrogate concepts, \xm{allowing for more precise control over the model's representation learning}.  
% \xm{Unfortunately, these approaches cannot guarantee the remaining concepts are sufficient preserved when the size of normal concepts become increased.} 

% \subsection{Neuron interpretation}
% \xm{In our work, we propose to address this problem from xxx perspective. }  
% \wenjie{do we need this section}

\section{Sparse Autoencoder for T2I Diffusion Models}

In this section, we introduce the training of an SAE for T2I diffusion model. We first discuss where to apply SAE for effective and efficient concept erasing in diffusion models in Section~\ref{ssec:where}. We then introduce how an SAE is trained in Section~\ref{ssec:sae_training}.

\subsection{Where to apply Spare Autoencoder.}
\label{ssec:where}
A T2I diffusion model comprises multiple modules that work jointly to generate an image. As a short preliminary, we take the Latent DM (LDM)~\cite{rombach2022high} as an example, wherein the text encoder and U-net are two main modules. The text encoder transforms input text prompts into embeddings $E$, which is used to guide the image generation process. The U-net module works as a noise predictor, taking the text embedding $E$, timestep $t$, and the noised latent representation $x_t$ as inputs to predict the noise added at time $t$. Specifically, with an initial noise $x_T \sim \mathcal{N}(\mathbf{0}, \mathbf{I})$, the generation process iteratively performs denoising operations on $x_T$, ultimately producing the final image $x_0$. Take DDPM~\cite{ho2020denoising} as an example, the denoise step can be formulated as:
\begin{equation}
\label{equ:ddim}
% \boldsymbol{x}_{t-1}=\sqrt{\alpha_{t-1}} \left(\frac{\boldsymbol{x}_t-\sqrt{1-\alpha_t} \epsilon_\theta\left(\boldsymbol{x}_t\right)}{\sqrt{\alpha_t}}\right)+\sqrt{1-\alpha_{t-1}-\sigma_t^2} \cdot \epsilon_\theta\left(\boldsymbol{x}_t\right)+\sigma_t \epsilon_t
    x_{t-1}=\frac{1}{\sqrt{\alpha_t}}\left(x_t-\frac{1-\alpha_t}{\sqrt{1-\bar{\alpha}_t}} \epsilon_\theta\left(x_t, E, t\right)\right)+\sigma_t \epsilon,
\end{equation}
where $\epsilon_\theta$ is realized by an U-net model, $\alpha_t, \bar{\alpha}_t, \sigma_t$ are pre-defined values and $\epsilon \sim \mathcal{N}(\mathbf{0}, \mathbf{I})$.

To disable the model's ability to generate unwanted images, numerous approaches propose modifying the U-net module to remove unwanted knowledge~\cite{zhang2024forget,kumari2023ablating,gandikota2023erasing,fan2023salun,huang2023receler,bui2024erasingadversarial,lu2024mace,li2024safegen,conceptpinpoint2025}. 
However, due to (1) the complexity of the U-Net architecture and (2) the iterative nature of the denoising process, which typically requires multiple steps (e.g., 50 in DDIM), even minor modifications to the U-Net can lead to unexpected outcomes, potentially degrading overall performance.

To mitigate the issue, we propose applying SAE to the text encoder to remove unwanted knowledge from the text embedding before feeding it into the U-Net. The key motivations behind this are as follows: 
\begin{packeditemize}
    \item During the image generation process, the text embedding $E$ plays a dominant role in encoding the semantic information to the generated images (cf. Equation~\ref{equ:ddim}). Consequently, erasing concepts at the text embedding level is sufficient to prevent their appearance in generated images.
    \item While previous works perform concept erasure within U-Net, their modifications mainly target text information processed within the CA layers (i.e., Key and Value projections, as detailed in the Appendix)~\cite{lu2024mace,conceptpinpoint2025}, which demonstrates the effectiveness of performing erasure on text embedding.
    \item Prior studies demonstrate that performing unlearning on the text encoder achieves the best robustness against adversarial attacks~\cite{zhang2024defensive}. 
\end{packeditemize}

% 1) during the image generation process, the text embedding $E$ plays a dominant role in encoding the semantic information to the generated images (cf. Equation~\ref{equ:ddim}). Consequently, perform erasing on text embedding is sufficient to erase target concepts within generated images 2) while previous works perform erasing within U-net, their erasure mainly focus on text information within CA layers (i.e., Key and Value, more details in Appendix)~\cite{lu2024mace,conceptpinpoint2025} and 3) prior studies demonstrate that performing adversarial unlearning on the text encoder achieves the best robustness against adversarial attacks~\cite{zhang2024defensive}. 
% Motivated by these observations, we propose applying SAE to the text encoder to remove unwanted knowledge from the text embedding before feeding it into the U-Net for concept erasure.
In the following section, we detail the training process of SAE.

\subsection{Training of SAE}
\label{ssec:sae_training}

\begin{figure}[tbp]
\centering
\setlength\tabcolsep{0.45pt}
\begin{tabular}{ccc}
    \subfloat{
        \includegraphics[width=0.4\textwidth,valign=c]{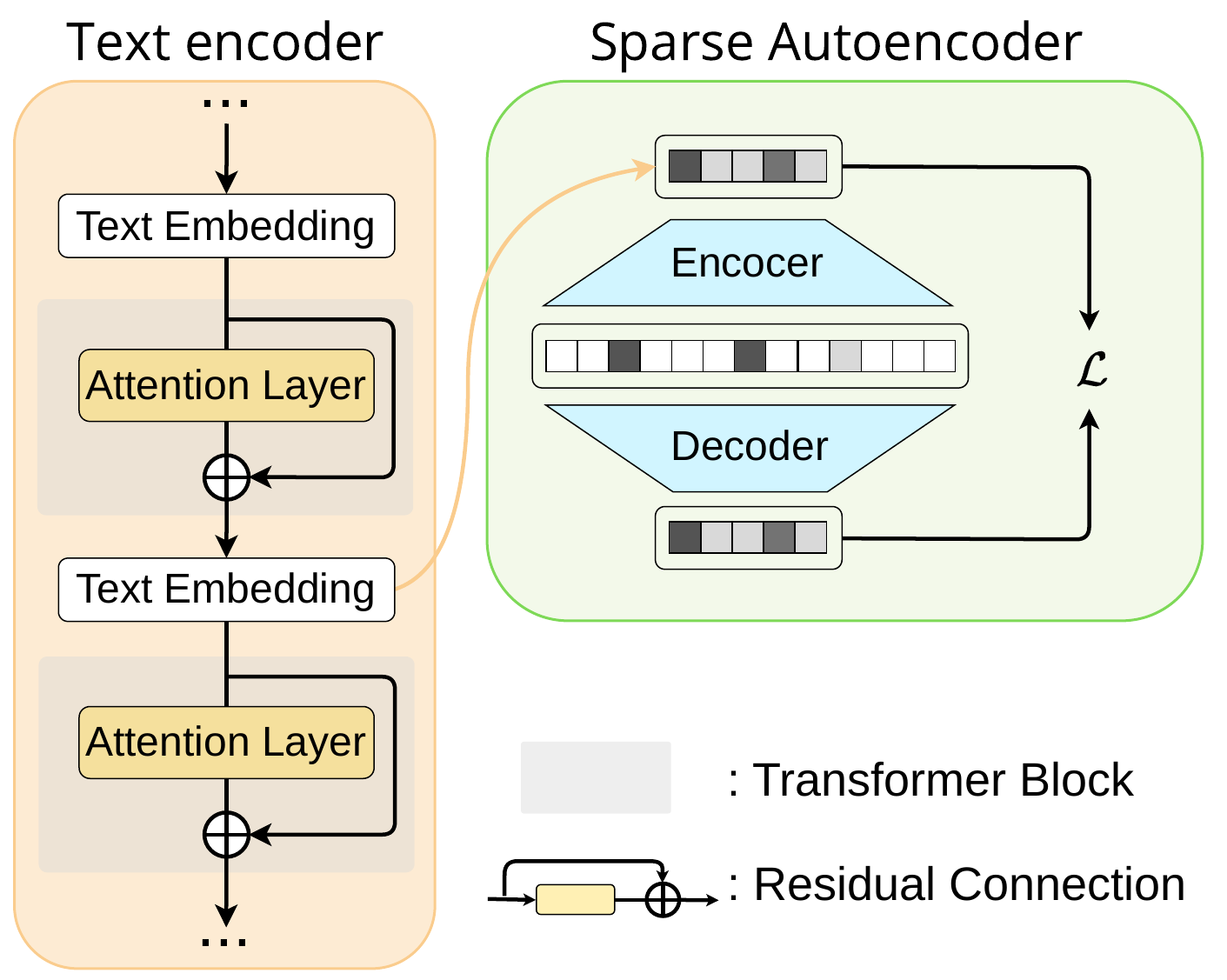}
    } 
\end{tabular}
\caption{Unsupervised training of SAE, which takes a token embedding obtained from the residual streamer in text encoder as an input and aims to reconstruct it with sparse features}
\label{fig:sae}
\end{figure}

The text encoder comprises a series of transformer blocks (cf. Figure~\ref{fig:sae}). Denote \(\mathcal{T}_l\) as the \(l\)-th transformer block, the text encoder with $L$ blocks can be roughly formulated as $\text{TextEncoder}= \mathcal{T}_L \circ ...\circ \mathcal{T}_1$.
To train an SAE for concept erasure, we focus on residual stream~\cite{elhage2021mathematical}, which is the output of a transformer block. 
The residual stream of \(l\)-th layer can be represented as:
\begin{equation}
    \textbf{e}_l = \mathcal{T}_l \circ ...\circ \mathcal{T}_1(\textbf{e}_0),\ \textbf{e}_l\in\mathbb{R}^{H\times d_{\text{in}}}
\end{equation}
where $\mathbf{e}_0$ is the embedding of the tokenized prompt, $H$ is the number of tokens composing the prompt, and $d_{\text{in}}$ is the output dimension. We assume that \(d_{\text{in}}\) are the same across different layers for notation simplicity. We train an SAE that aims to learn the sparse features for each token embedding $\textbf{e}_l^h, h\in[1,.., H]$. Therefore, for a prompt with $H$ tokens, we get $H$ samples to train SAE.

In our work, we adopt K-sparse autoencoder (K-SAE)~\cite{makhzani2013k}, which could explicitly control the number of active latents by only keeping the $K$ largest activations and zeros the rest for reconstruction.
Let $\textbf{e}\in \mathbb{R}^{d_{\text{in}}}$ refers to a single SAE training example. The encoder and decoder within K-SAE are then defined as follows:
\begin{equation}
\begin{aligned}
    \textbf{z}&=\operatorname{TopK}\left(W_{\text{enc}}\left(\textbf{e}-\textbf{b}_{pre }\right)\right)\\
    \hat{\textbf{e}} &= W_{\text{dec}}\textbf{z} + \textbf{b}_{\text{pre}}
\end{aligned}
\end{equation}

where $W_{\text{enc}}\in\mathbb{R}^{d_{\text{in}} \times d_{\text{hid}}}$ and $W_{\text{dec}}\in\mathbb{R}^{d_{\text{hid}} \times d_{\text{in}}}$ are learned matrix of encoder and decoder, $b_{pre}$ is the bias term. $d_{\text{hid}}$ is significantly larger than $d_{\text{in}}$ to enforce the sparsity of learned  features.

The training objective is:
\begin{equation}
\mathcal{L}(\mathbf{x})=\|\mathbf{x}-\hat{\mathbf{x}}\|_2^2+\alpha\mathcal{L}_{aux},
\end{equation}
where $\mathcal{L}_{aux}$ is the reconstruction error using top $K_{aux}$ ($K_{aux}>K$) feature activations, to prevent dead features that have not been fired on a large number of training samples~\cite{gao2024scalingevaluatingsparseautoencoders}, scaled with coefficient $\alpha$. 

% By optimizing the MSE loss between $e_l[h]$ and $\hat{e}_l[h]$ as well as $\mathcal{L}_{aux}$ (cf. Section~\ref{ssec:sae}), it encourages that the reconstruction be a sparse linear combination of the sparse features. 

We refer to the activation of the $\rho$-th learned feature as $z^\rho\in\mathbb{R}$. Its associated feature vector $f_\rho \in \mathbb{R}^{d_{in}}$ is a column in the decoder matrix $W_{\text{dec}}=\left(\mathbf{f}_1|\cdots| \mathbf{f}_{n_f}\right) \in \mathbb{R}^{d_{\text{hid}} \times d_{\text{in}}}$. As a result, we can represent each token embedding as a sparse sum 
\begin{equation}
    \textbf{e} \approx  \sum_{\rho=1}^{d_{hid}} z^\rho \mathbf{f}_\rho,
    \text { with } ||\textbf{z}||_0 \leq K.
\end{equation}
The details of training SAE are presented in Appendix.

% For more details on the SAE training
% and for training metrics consider the App.\todo{Training SAE}.

% \section{Methodology}

\section{Methodology}
\label{sec:method}
\begin{figure*}[htbp]
\centering
\setlength\tabcolsep{0.5pt}
\begin{tabular}{ccc}
    \subfloat{
        \includegraphics[width=1\textwidth,valign=c]{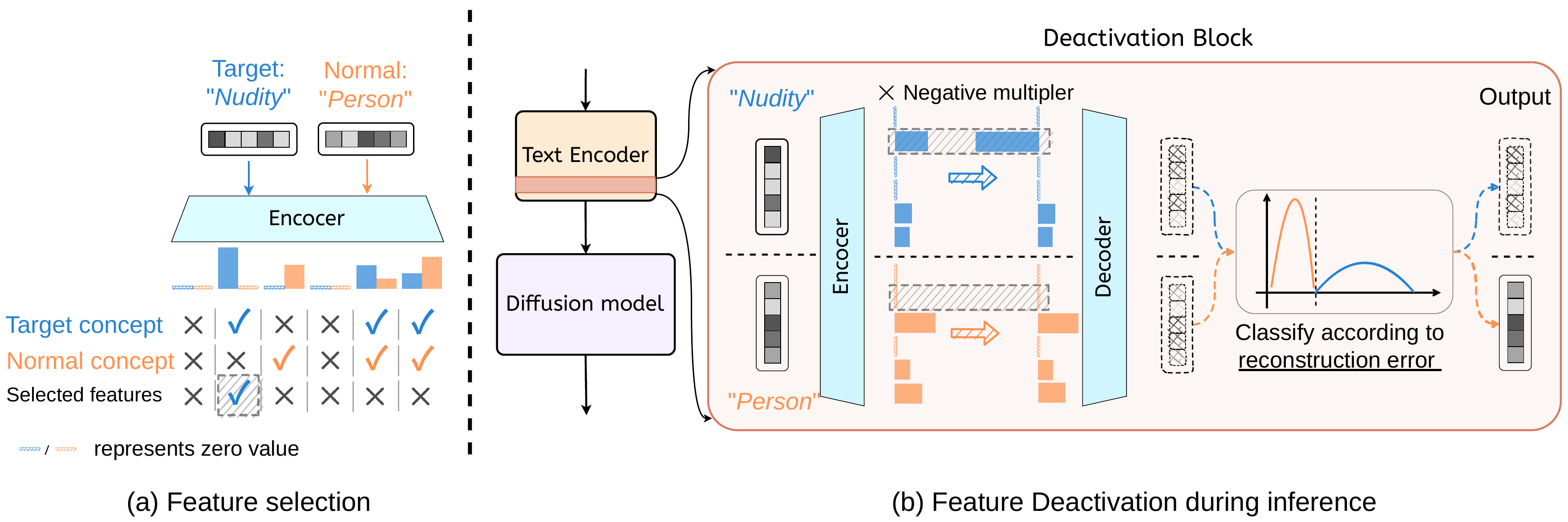}
    } 
\end{tabular}
\caption{(a) With a well-trained SAE, we identify unique features of target concepts by contrast with normal concepts; (b) we wrap SAE as a deactivation block and insert it into the text encoder for concept eraser.}
\label{fig:sae_block}
\end{figure*}

% learning to represent the data efficiently with a sparse set of features, enabling it to capture important, low-dimensional structures
% :Identifying Unique Features for Concept Erasure in Sparse Autoencoders

%: Erasing Features and Integrating Sparse Autoencoders into Text Encoders for Concept Manipulation

In this section, we introduce \name, a framework to erase multiple concepts in a pretrained T2I diffusion model. 

\subsection{Overview}
With a well-trained SAE, a straightforward approach to erasing unwanted knowledge within the text embedding is to deactivate features associated with the target concepts during the inference of the text encoder. However, target concepts often share certain features learned by SAE with normal concepts (e.g., \textit{"nudity"} and \textit{"person"} both contain information related to the \textit{"human body"}). As a result, indiscriminately deactivating all related features could affect the generation of normal concepts.

To solve the problem, we propose a simple yet effective contrast-based method to identify features specific to the target concepts (Section~\ref{ssec:feature_selection}). By only deactivating carefully selected features, \name could effectively remove unwanted knowledge while having limited effects on normal concepts. Furthermore, we find that SAE can be exploited as a zero-shot classifier to distinguish between target and remaining concepts. This enables selective concept erasure, further reducing the impact on normal concepts (Section~\ref{ssec:concept_erasing}). Figure~\ref{fig:sae_block} depicts the workflow of \name.
% mainly involves three steps: 1) feature selection, which aims to identify features specific to the target concepts by utilizing a simple yet effective contrast approach, 2) feature deactivation, during the inference of the text encoder, we encode each embedding with SAE, deactivate selected features, and decode it back, and 3) 
Overall, it is designed to meet the following criteria:
\begin{packeditemize}
% \item{Efficacy:} 
\item{Effectiveness}: The unlearned model could not generate images of target concepts even when prompted with texts related to target concepts. 

\item{Robustness:} 
The model should also prevent the generation of images that are semantically related to synonyms of the targeted concepts, ensuring that erasure is not restricted to exact prompt wording.

\item{Specificity:} The erasure should target only the specified concepts, with minimal or no impact on the remaining concepts.  

\item{Expandability:} 
When erasing new concepts, the algorithm can be easily extended without the need for additional training.

\end{packeditemize}

% To fulfill the above requirement, we introduce Interpret-then-deactivate (\name), a framework that enables the four targets above. We first show how to apply SAE in a T2I diffusion model. With a well-trained SAE, we then present how to construct the deactivation block to enable concept erasing.

% which decomposes each token embedding into a linear combination of spare features, we can erase the knowledge of target concepts by simply deactivating the features that the concept activates. 
% One challenge is how to discover the set of features that are specific to the target concepts. 
% With a well-trained SAE, each token embedding can be represented as a linear combination of sparse features, allowing us to erase target concept knowledge by deactivating the relevant features. The key challenge lies in identifying the set of features specific to the target concepts.

\subsection{Feature Selection.}
\label{ssec:feature_selection}
% Given a target concept $C$, which is usually composed of multiple tokens, we denote them as $C_1, ..., C_H$. For each token embedding, SAE encodes it into a linear combination of sparse features. As different token embedding activates different sets of sparse features, to fully capture the features that the concept activates, we select the top $k$ features that can be activated by any token embedding of the concept according to $s^\rho$.

\paragraph{Select Features for a Concept.}
A concept \( C \) is typically composed of multiple tokens, denoted as \( C_1, \ldots, C_H \) where \( H \geq 1 \). For example, "Bill Clinton" consists of the tokens "Bill" and "Clinton". SAE decomposes the embedding of each token into a set of features. To select features that can represent the concept, we collect features from all token embedding and select the top \( K_{\text{sel}} \) features based on their activation values \( s^\rho \). 
% According to the autoregressive property of the text encoder~\cite{vaswani2017attention}, the embedding of the last token encapsulates the information of the entire concept.  
% One possible approach is to select features decomposed from the last token embedding to represent the concept. However, our experiments show that this method may underestimate certain features, especially when concepts consist of multiple tokens. 
% To address this, we aggregate features decomposed from all token embeddings to represent the concept. We then select the top \( K_{\text{sel}} \) features based on their activation values \( s^\rho \).
% These features are chosen based on their potential to be activated by any token embedding of the concept, as determined by the criterion \( s^\rho \).
Formally, we denote $F$ as the set of indices of features specific to the concept $C$, 
\begin{equation}
\begin{aligned}
    F = \{\rho |s_C^\rho\in \operatorname{TopK}(s_C^1,...,s_C^{d_{hid}})\}, \\ \text{where } s_C^\rho = \operatorname{max}(s_1^\rho,...,s_H^\rho).
\end{aligned}
\end{equation}
% In the following context, we mix the use of text embedding and token embedding. Applying SAE to a text embedding refers to applying it to each token embedding it contains.

\paragraph{Select Features for Concept Erasing.}
After selecting features associated with target concepts, we propose using the features that are specifically activated by the target concepts for erasure. To achieve that, we suggest a simple yet effective contrast-based approach. Specifically, given a retain set $\mathcal{C}_{\text{retain}}$ comprising normal concepts, the features specific to the target concept (i.e. feature to deactivate) $\hat{F}_{\text{tar}}$ can be found by eliminating features that can be activated by normal concepts in $\mathcal{C}_{\text{retain}}$:
\begin{equation}
    \hat{F}_{\text {tar }}=F_{\text {tar }} \backslash \bigcup_{C_r \in C_{\text {retain }}} F_{C_r}.
\end{equation}

In our experiments, we use the concepts employed for the utility-preserving regularization term in previous works~\cite{kumari2023ablating,zhang2024defensive,fan2023salun,lu2024mace} as the retain set for comparison. Notably, our approach does not require fine-tuning on these concepts, thereby avoiding the introduction of new biases and making more effective use of them. 

When erasing multiple concepts, we select features for
erasing as the union of the specific features associated with each target concept.
Specifically, let $\mathcal{C}_{\text{tar}}$ denote the set of target concepts.   
\begin{equation}
F_{\text{erase}} = \bigcup_{C \in \mathcal{C}_{\text{tar}}} \hat{F}_C.
\end{equation}

\subsection{Concept Erasing}
\label{ssec:concept_erasing}

To erase knowledge of target concepts within the text embedding, we first encode the text embeddings into their activations \( s \) using SAE. We then modify each activation component as follows:  
\begin{equation}
\label{equ:deactivation}
\hat{s}^\rho = 
\begin{cases}
s^\rho \cdot \tau, & \text{if } \rho \in F_{\text{erase}} \\ 
s^\rho, & \text{otherwise}
\end{cases}
\end{equation}
where \( F_{\text{erase}} \) represents the set of features selected before, and \( \tau \) is a scaling factor controlling the degree of deactivation. Finally, the decoder reconstructs the modified embedding as \( \hat{\mathbf{e}} = \mathsf{Dec}(\hat{s}) \).

To build an unlearned model, we wrap SAE as a deactivation block inserted into the intermediate layers of the text encoder (Figure~\ref{fig:sae_block} (b)). Any text embedding inputted to the block would erase the information of target concepts. As a result, the generated image guided by the text embedding would not contain target concepts. The block is plug-and-play, which means we do not need to
fine-tune the diffusion model, making it highly efficient.
Moreover, since the selected features are not activated by normal concepts, Equation~\ref{equ:deactivation} does not influence normal concepts.

\paragraph{Selective feature deactivation}
The inherent reconstruction loss of SAE may still influence the generation of normal concepts.
To alleviate the issue, we propose a mechanism to selectively apply SAE for concept erasure.
Given a text embedding \( \mathbf{e} \) and its reconstructed version \( \hat{\mathbf{e}} \), we construct a classifier $G$ to identify whether \( \mathbf{e} \) contains information about target concepts. The classification is based on the reconstruction loss between \( \mathbf{e} \) and \( \hat{\mathbf{e}} \):
\begin{equation}
\label{equ:deactivation}
 G(\mathbf{e})= 
\begin{cases}
1, & \text{if } ||\mathbf{e}-\hat{\mathbf{e}}||^2 <\tau \\ 
0, & \text{if } ||\mathbf{e}-\hat{\mathbf{e}}||^2 \geq \tau
\end{cases}
\end{equation}
where $\tau$ is the threshold.
Finally, the deactivation block outputs \( \mathbf{e} \) for subsequent computation if it does not contain target concept information; otherwise, it outputs \( \hat{\mathbf{e}} \).

% Our observation lies in that the target concepts are more sensitive to the selected unique features. 
% As a result, suppressing these features would significantly change the reconstructed embedding, inducing a high reconstruction loss. On the contrary, the remaining concepts do not comprise the selected features, hence the reconstruction loss is mainly induced by the intrinsic loss of SAE. 

\begin{figure}[htbp]
\centering
\setlength\tabcolsep{0.6 pt}
% \begin{tabular}{ccc}
    \subfloat{
        \includegraphics[width=0.45\textwidth,valign=c]{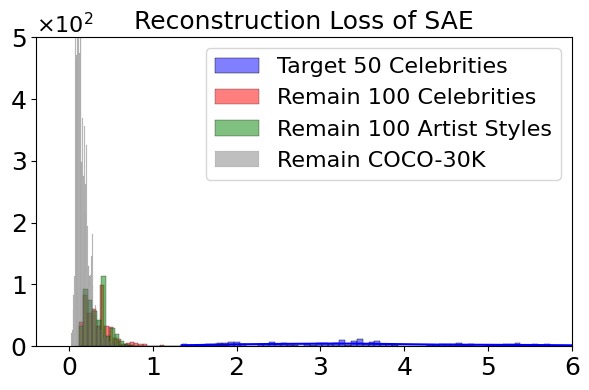}
    } 
% \end{tabular}
\caption{Histogram of reconstruction MSE loss on target concepts and remaining concepts.}
\label{fig:mse}
\end{figure}

Figure~\ref{fig:mse} demonstrates the effectiveness of the classifier. We select 50 celebrities as target concepts for erasure. For the remaining concepts, we include 100 different celebrities, 100 artistic styles, and the COCO-30K dataset. It shows that there is a clear boundary in the reconstruction loss between the target concepts and the remaining concepts.

% We have the following assumption:
% In summary, our assumption is \textit{The reconstruction error of target concepts induced after feature suppression is more significant than the remaining concepts}. 

% \begin{assumption}
%     The reconstruction error of target concepts induced after feature suppression is more significant than the remaining concepts:
% \end{assumption}

% \wenjie{why present this assumption here, is it used later?}
% \wenjie{shall we also formally define other assumptions previously used}

% \section{Methodology}

\begin{figure*}[htbp]
\centering
\setlength\tabcolsep{0.5pt}
\begin{tabular}{ccc}
    \subfloat{
        \includegraphics[width=0.95\textwidth]{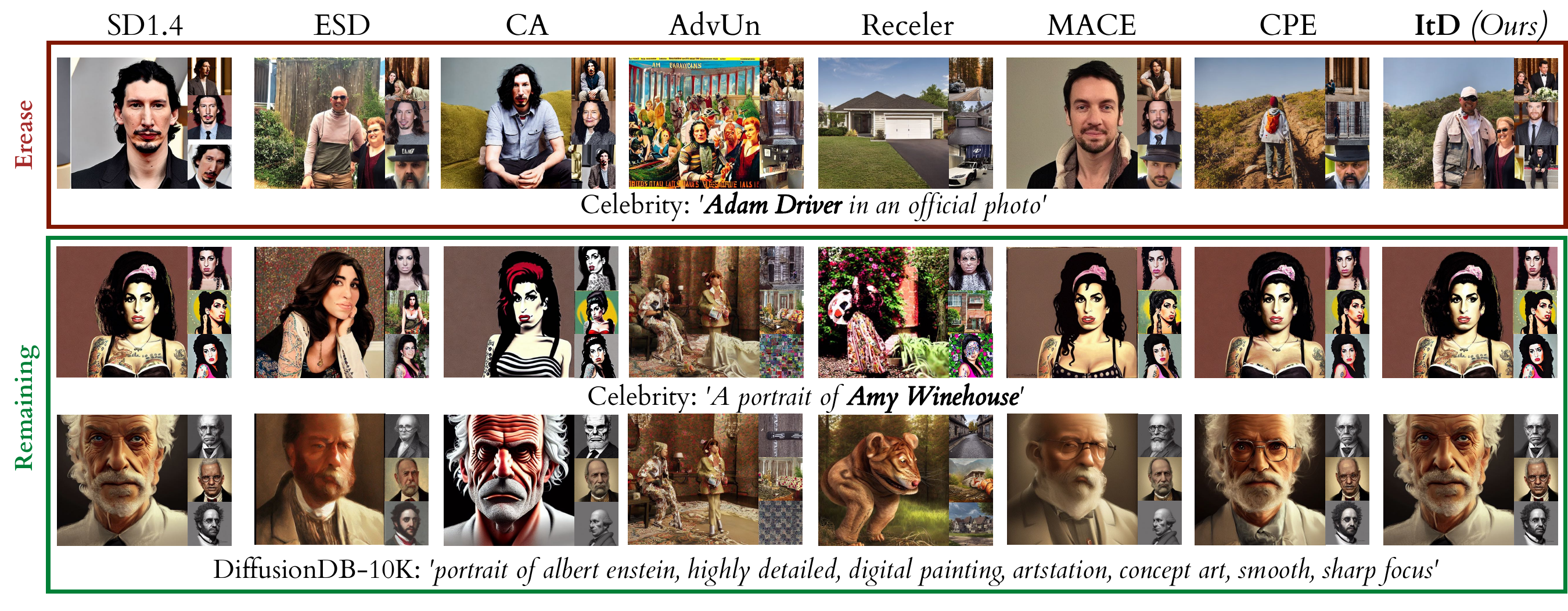}
    } 
\end{tabular}
\caption{Qualitative results of our \name and baselines on multiple concepts erasing. We erased 50 celebrities at once. The remaining celebrity concepts serve as surrogate concepts in the baselines and as training data for SAE in \name, whereas the concepts in DiffusionDB-10K are used solely during the generation process.}
\label{fig:celebrity_erase}
\end{figure*}

\begin{table*}[ht]
\centering
\caption{Quantitative results on celebrities erasure. We use CLIP Score (CS) and GCD accuracy (ACC) for target celebrities. We measured CS and FID for COCO-30K and DiffusionDB-10K, or KID for the other remaining concepts.}
\label{tab:erase_celeb}
\resizebox{\textwidth}{!}{
\begin{tabular}{cccccccccccc}
\toprule
\multirow{3}{*}{Methods} & \multicolumn{2}{c}{Target Concepts} & \multicolumn{7}{c}{Remaining Concepts} & \multicolumn{2}{c}{Unpredictable Concepts}  \\ 
\cmidrule(lr){2-3} \cmidrule(lr){4-10} \cmidrule(lr){11-12}
      & \multicolumn{2}{c}{50 Celebrities} & \multicolumn{3}{c}{100 Celebrities} & \multicolumn{2}{c}{100 Artistic Styles} & \multicolumn{2}{c}{COCO-30K} & \multicolumn{2}{c}{DiffusionDB-10K} \\ 
\cmidrule(lr){2-3} \cmidrule(lr){4-10} \cmidrule(lr){11-12}
      & CS $\downarrow$ & ACC$\%$ $\downarrow$ & CS $\uparrow$ & ACC$\%$ $\uparrow$ & KID($\times 100$) $\downarrow$ & CS $\uparrow$ & KID($\times 100$) $\downarrow$ & CS $\uparrow$ & FID($\times 100$) $\downarrow$ & CS $\uparrow$ & FID $\downarrow$ \\
\midrule
ESD-x~\cite{gandikota2023erasing}   & 24.41 &  7.30  & 26.23 & 10.39 & 2.66 & 28.23 & 0.01 & 29.55 & 14.40 & 28.94  &  9.46 \\ 
AC~\cite{kumari2023ablating}        & 33.63 &  90.16 & 33.98 & 87.04 & 1.86 & 28.47 &  0.38  & 30.91    &  16.91  & 31.28    &  8.55  \\
AdvUn~\cite{zhang2024defensive}     & 16.71 &  0.00  & 17.61 & 2.84 & 14.80 & 19.29 &  10.29  & 18.25 &  47.38  &  14.53   & 63.61   \\
Receler~\cite{huang2023receler}     & \textbf{12.84} &  0.00  & 13.37 & 86.45 & 18.09 & 24.80 &  5.23  & 29.98 &  15.85   &  27.85   & 18.96  \\
MACE~\cite{lu2024mace}              & 24.60 & 3.29 & 34.39 & 84.64 & 0.23  &  27.75   &  0.47  &  30.38   &  \textbf{14.03}   & 29.59    & 11.10  \\ 
CPE~\cite{conceptpinpoint2025}  & 20.79 &  0.37&   34.82 &  88.26 &0.08 & 29.01 &   0.01   &  30.86   & 14.62 &  31.84 &  4.18   \\ 
\textbf{ItD} (Ours)         & 19.65 & \textbf{0.00} & \textbf{34.87} & \textbf{89.12} &  \textbf{0}  & \textbf{29.02} &  \textbf{0} &  \textbf{31.02}    &  14.72  & \textbf{32.06}   & \textbf{0.91} \\ 
\midrule
SD1.4 \cite{rombach2022high}  & 34.49 &  90.56  & 34.87 & 89.12 & - &  29.02  &  -  &  31.02  & 14.73 & 32.17  &  -  \\ 
\bottomrule
\end{tabular}
}
\end{table*}

\section{Experiments}

In this section, we conduct comprehensive experiments to evaluate the effectiveness of our approach.
% We are keen to evaluate the following questions: 1) How is the generation capacity on remaining concepts in our approaches compared with state-of-the-art baselines in celebrity erasing, style erasing and nudity concept erasing. 2) How the generation capacity on unpredictable concepts manifests, referring to those not present in the training dataset of the SAE model. 3) How the robustness of our model manifests in the NSFW domain pertains to the probability of generating nudity images when subjected to adversarial attacks.

\paragraph{Setting.} 
% We experiment upon multiple concept erasing tasks: celebrities erasure and artistic styles erasure. 
% To evaluate the generation capacity on remaining concepts,  
We consider four domains for concept erasing tasks: celebrities, artistic styles, COCO-30K captions, and DiffusionDB-10K captions, where the last dataset contains 10K captions collected from DiffusionDB~\cite{DiffusionDB}. These captions are constructed by collecting chat messages from Stable Diffusion Discord channels, representing the user prompts in a real-world setting. Next, we conduct experiments on the removal of explicit contents and evaluate the efficacy on I2P dataset~\cite{schramowski2023safe}. We also evaluate the robustness against adversarial prompts using the red-teaming tools Unlearning-Diff~\cite{zhang2025generate}.
We generate images using SD1.4 ~\cite{rombach2022high} with DDIM in 50 steps. 

\paragraph{SAE Training Setting}
For celebrity and artistic style erasure, we train an SAE using 200 celebrity names and 200 artist names from MACE~\cite{lu2024mace}. Additionally, we incorporate the full captions from COCO-30K. The celebrity and artist names are embedded into 100 and 35 prompt templates, respectively, following the setup in CPE~\cite{conceptpinpoint2025}. This results in a total of 57,000 text samples for training the SAE.  
For nudity erasure, we collect 10,000 captions from DiffusionDB~\cite{DiffusionDB} with an NSFW score above 0.8, along with all captions from COCO-30K, to train the SAE. Further details on the training process can be found in the Appendix~\ref{sec:sae_training}.

% When performing the Celebrity Erasure and Style Erasure, we use a pre-trained SAE model, eliminating the need to train separately for each concept or fine-tune the Stable Diffusion model. We conduct the training on the block \textit{text\_model.encoder.layers.8}; the input dimension of our model is 768. To better capture the characteristics of the prompt, we sparsely map this 768-dimensional input to a 524,288-dimensional space. We set the training prompts from the celebrity, style, COCO-30K datasets, running the training for 6,000 epochs. We set TopK to 64.

% For the Explicit Content Erasure task, we used the Diffusiondb\_nsfw and COCO datasets and trained for 4,000 epochs, applying the same TopK setting and block configuration. Diffusiondb\_nsfw is a subset of DiffusionDB, specifically extracted to include only the NSFW(Not Safe for Work)-related prompts.

% We generate \xm{xx} images for target concepts,  prompted by ``a photo of
% the \{erased class name\}'', and generate \xm{xx} images for the remaining concepts, prompted by ``a photo of the \{normal class name\}''. 

% \xm{of xxx} and

\paragraph{Baselines.} We compare with six baselines including four fine-tuning-based approaches: ESD~\cite{gandikota2023erasing}, AC~\cite{kumari2023ablating}, AdvUn~\cite{zhang2024defensive}, Receler~\cite{huang2023receler}.
and two inference-based approaches:
MACE~\cite{lu2024mace}, and CPE~\cite{conceptpinpoint2025}. The implementation details are provided in Appendix~\ref{sec:preliminary}.

\paragraph{Metrics.} 
% To evaluate the quality of generated images for both target and remaining concepts, 
We adopt CLIP score{~\cite{hessel2021clipscore}} to evaluate the quality of generated images, where a lower score indicates an effective erasure for target concepts, and a higher score refers to better preservation for the remaining concepts. 
% For target concepts, a lower CLIP score represents more effective erasure and for remaining concepts, a higher CLIP score represents better preservation. 
Additionally, for celebrity erasure experiments, we utilize the GIPHY Celebrity Detector~\cite{celebritydetector} to measure the top-1 accuracy (\textbf{ACC}) of generated celebrity images. A lower accuracy is better for target concepts, and a high accuracy is better for remaining concepts. For COCO-30K and DiffusionDB-10K captions, we evaluate their Frechet Inception Distance (\textbf{FID})~\cite{heusel2017gans}. A lower value is better. Following~\cite{conceptpinpoint2025}, we evaluate the Kernel Inception Distance (\textbf{KID}) for other remaining concepts, which is more stable and reliable for a smaller number of images.

\subsection{Celebrity Erasure}~\label{ssec:celeb_erasure}  
We select 50 celebrities as the target concepts from the list of celebrities provided by~\cite{lu2024mace}, which consists of 200 celebrities.
For the remaining concepts, we consider two domains: 100 celebrities and 100 artistic styles.
We generated 25 images using 5 prompt templates with 5 random seeds, resulting in 2,500 images for each remaining domain. We also use COCO-30K and DiffusionDB-10K as remaining concepts. Notably, the captions in DiffusionDB-10K are not used for training SAE, allowing us to evaluate the effectiveness of our approach on unpredictable concepts. 
% The threshold $\tau$ is selected with 3 different templates for all target celebrities and is set as a minimal reconstruction loss.

Figure~\ref{fig:celebrity_erase} presents the qualitative results of erasing multiple celebrities. The results indicate that \name and all baselines, except CA, effectively remove the target concepts. CA struggles in this scenario due to the large number of target concepts, exceeding its capacity for erasure.  
For the remaining concepts, AdvUn and Receler cause significant degradation in image quality, likely because adversarial unlearning introduces broader disruptions to the generation process. In contrast, MACE and CPE preserve the quality of remaining celebrity images, as they are trained within the same domain. However, images from DiffusionDB-10K exhibit noticeable deviations from those generated by the original model. Among all methods, \name is the only approach that maintains the same generation quality as the original model.  

Table~\ref{tab:erase_celeb} presents quantitative results on erasing target concepts while evaluating the impact on remaining concepts. The results demonstrate that \name has the least effect on remaining concepts and outperforms all baselines by a large margin in preserving them, while still achieving strong erasure performance.

\begin{figure*}[ht]
\centering
\setlength\tabcolsep{0.5pt}
\begin{tabular}{ccc}
    \subfloat{
        \includegraphics[width=0.95\textwidth,valign=c]{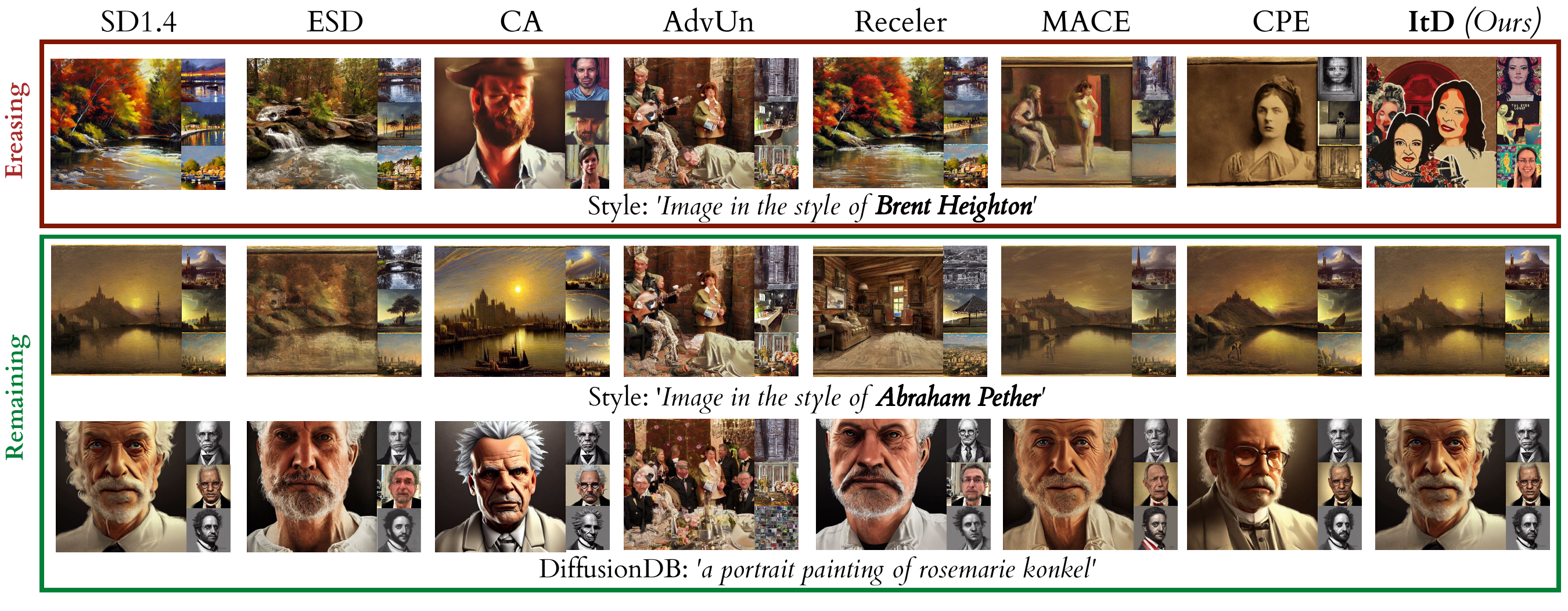}
    } 
\end{tabular}
\caption{Qualitative results of our \name and baselines on multiple concepts erasing. We erased 100 artist styles at once. The remaining artist style concepts serve as surrogate concepts in the baselines and as training data for SAE in \name, whereas the concepts in DiffusionDB-10K are used solely during the generation process.}
\label{fig:style_erase}
\end{figure*}

\begin{table*}[t]
\centering
\footnotesize
\caption{Quantitative results on artistic styles erasure. We used CLIP Score (CS)for target artistic styles. We measured CS and FID for COCO-30K and DiffusionDB-10K, or KID for the other remaining concepts.}
\label{tab:erase_artistic}
\resizebox{\textwidth}{!}{
\begin{tabular}{ccccccccccc}
\toprule
\multirow{3}{*}{Methods} & Target Concepts & \multicolumn{7}{c}{Remaining Concepts}  & \multicolumn{2}{c}{Unpredictable Concepts}  \\ 
\cmidrule(lr){2-2} \cmidrule(lr){3-9} \cmidrule(lr){10-11}
      & 100 Artistic Styles & \multicolumn{2}{c}{100 Artistic Styles} & \multicolumn{3}{c}{100 Celebrities} & \multicolumn{2}{c}{COCO-30K} & \multicolumn{2}{c}{DiffusionDB-10K} \\ 
\cmidrule(lr){2-2} \cmidrule(lr){3-9} \cmidrule(lr){10-11}
      & CS $\downarrow$  & CS $\uparrow$ & KID($\times 100$) $\downarrow$ & CS $\uparrow$ & ACC$\%$ $\uparrow$ & KID($\times 100$) $\downarrow$ & CS $\uparrow$ & FID($\times 100$) $\downarrow$  & CS $\uparrow$ & FID $\downarrow$ \\
\midrule
ESD-x~\cite{gandikota2023erasing} & 20.89 & 28.90 & 0.65 & 30.42& 81.86   & 0.81    & 29.52 &  15.19 & 28.11 & 11.54     \\ 
AC~\cite{kumari2023ablating}      & 28.91 & 28.33 & 1.25 & 34.77& \textbf{93.71}   & 0.25   & 30.97  & 16.19 & 31.21 &10.20     \\
AdvUn~\cite{zhang2024defensive}   & \textbf{18.94} & 19.28 & 9.65 & 17.78& 0.0  & 13.90    & 18.11   & 43.24  & 12.63 & 62.84    \\
Receler~\cite{huang2023receler}   & 24.90 & 24.96 & 2.11 & 32.76& 86.68 & 0.44   & 29.29  & 16.25  & 26.83 & 22.00  \\
MACE~\cite{lu2024mace}            & 22.59 & 28.95 & 0.25 & 26.87& 10.79 & 0.25  & 29.51 & \textbf{12.71} & 25.77 & 16.70 \\ 
CPE~\cite{conceptpinpoint2025} & 20.67  & 28.95 &0.01 &   34.81   &  89.80   &  0.04  & 30.95 & 14.77   &  31.60  & 5.72 \\ 
\textbf{ItD} (Ours)                        & 19.88 & \textbf{29.02} & \textbf{0.00} & \textbf{34.87} & 89.12 & \textbf{0.00}  & \textbf{31.02}   & 14.71  &  \textbf{31.91} & \textbf{1.07}   \\ \midrule
SD1.4~\cite{rombach2022high}      & 29.72 & 29.02 & - & 34.87 & 89.12 &  -   &  31.02 &  14.73 & 32.17 &-  \\ 
\bottomrule
\end{tabular}           
}
\end{table*}

\begin{table}
\footnotesize 
    \centering
    \caption{Robust concept erasure against adversarial attack: UnlearnDiff~\cite{zhang2025generate}}
    \label{tab:robust}
    % \resizebox{0.5\textwidth}{!}{
    \begin{tabular}{c|c}
    \toprule
    Method & ASR ($\downarrow$) \\
    \midrule
    FMN\cite{zhang2024forget} & 97.89 \\
    ESD~\cite{gandikota2023erasing} & 76.05 \\
    UCE~\cite{gandikota2024unified} & 79.58 \\
    MACE~\cite{lu2024mace} & 66.90\\
    AdvUnlearn~\cite{zhang2024defensive} & 21.13 \\
    % CPE (Ours) & $\underline{1.14}$ \\
    \name (Ours) & \textbf{12.61} \\
    \bottomrule 
    \end{tabular}
    % }
\end{table}

\begin{table*}
\centering
\caption{Results of the detected number of explicit contents using NudeNet detector on I2P and preservation performance on COCO 30K with CS, FID.}
\label{tab:erase_nudity}
\resizebox{\textwidth}{!}{
\begin{tabular}{cccccccccccc}
\toprule
\multirow{2}{*}{Methods}  & \multicolumn{9}{c}{Number of nudity detected on I2P (Detected Quantity)} & \multicolumn{2}{c}{COCO-30K}  \\ 
\cmidrule(lr){2-10} \cmidrule(lr){11-12}
    & Armpits & Belly & Buttocks & Feet & Breasts (F) &  Genitalia (F)&  Breasts (M) & Genitalia (M)& Total& CS $\uparrow$  & FID $\downarrow$ \\

\midrule
FMN~\cite{zhang2024forget} & 43 & 117 & 12 & 59 & 155 & 17 & 19 & 2 & 424 & 30.39 & 13.52\\
ESD-x~\cite{gandikota2023erasing}  &  59 & 73 & 12 & 39 & 100 & 4 & 30 & 8 & 315 & 30.69 & 14.41  \\ 
AC~\cite{kumari2023ablating}  &153 & 180 & 45 & 66 & 298 & 22 & 67 & 7 & 838 &  \textbf{31.37} & 16.25    \\
AdvUn~\cite{zhang2024defensive}  & 8   &  \textbf{0}  & \textbf{0} & 13 &  1 & 1 &  \textbf{0} & \textbf{0} &  28& 28.14  & 17.18 \\
Receler~\cite{huang2023receler}& 48  &  32 & 3 & 35 & 20 & 0 & 17 & 5 & 160 & 30.49 & 15.32\\
MACE~\cite{lu2024mace}   & 17 & 19 & 2 & 39 & 16 & 0 & 9 & 7 & 111 & 29.41 & \textbf{13.40}\\ 
CPE~\cite{conceptpinpoint2025}  & 10 & 8 & 2 & 8 & 6  & 1 & 3 & 2 &  40  & 31.19 & 13.89 \\ 
\textbf{ItD} (Ours)   & \textbf{0} &  2  & 3 & \textbf{3} & \textbf{0} & \textbf{0} &  \textbf{0}&10 & \textbf{18} & 30.42 & 14.64 \\ \midrule
SD v1.4~\cite{rombach2022high}  & 148 & 170 & 29 & 63 & 266 & 18 & 42 & 7 & 73 &31.02 & 14.73 \\
SD v2.1~\cite{sd2}  & 105 & 159 & 17 & 60 & 177 & 9 & 57 & 2 & 586 & 31.53 & 14.87 \\
\bottomrule
\end{tabular}
}
\end{table*}

\subsection{Artistic Styles Erasure}
We select 100 artistic styles as the target concepts following~\cite{lu2024mace} and utilize the same remaining concepts in Section~\ref{ssec:celeb_erasure}, including 100 celebrity,  100 artistic styles, COCO-30K, and DiffusionDB-10K. We utilize the same SAE used for celebrity erasure. 
% To select features specific to 100 target artistic styles, we contrast them with the remaining 100 artistic styles. 
Figure~\ref{fig:style_erase} shows the qualitative results of erasing multiple artist styles. Among all baselines, Receler fails to erase the target artistic style, possibly because it adopts a mask to identify pixels for erasure within the image, which fails when erasing styles, leading to unsuccessful erasure.
Table~\ref{tab:erase_artistic} shows results for artistic style erasure that are consistent with those from celebrity erasure. \name effectively distinguishes between target artistic styles, remaining artistic styles, and celebrities, ensuring strong erasure performance while preventing degradation of remaining concepts. Additionally, it has the least impact on DiffusionDB-10K and outperforms all baselines by a large margin.

\subsection{Explicit Content Erasure}
We evaluate the effectiveness of erasing explicit contents on the I2P dataset~\cite{schramowski2023safe}. It consists of 4,703 prompts without inappropriate words but would generate explicit images with stable diffusion. To erase explicit concepts, we adopt four keywords as target concepts to select features following~\cite{lu2024mace}: 'nudity', ‘naked’, ‘erotic’, and ‘sexual’. We employ the NudeNet detector~\cite{NudeNet} to measure the frequency of explicit content. The threshold is set to $0.6$~\cite{lu2024mace}. For preservation performance evaluation, we utilize COCO-30K. We train an SAE using the captions of COCO-30K. 
% Additionally, to learn the features of inappropriate words, we collect 10K prompts from DiffusionDB~\cite{DiffusionDB} with an NSFW score larger than 0.8. 
Table~\ref{tab:erase_nudity} shows the number of explicit contents detected by the NudeNet detector. The results show that \name resulted in the fewest detected explicit contents compared with baselines.

\subsection{Robustness}
To evaluate the robustness against the adversarial attack prompts, we utilize UnlearnDiff~\cite{zhang2025generate} as the red-teaming tool to verify the robustness of our \name. We evaluate on I2P dataset and report the attack success rate (ASR) as the evaluation metric to measure the ratio of generated images containing explicit content. The results are shown in Table~\ref{tab:robust}. 

From Table~\ref{tab:robust}, we can find that \name demonstrates robust erasure of target concepts competitive to recent robust
methods AdvUnlearn ~\cite{zhang2024defensive} and MACE ~\cite{lu2024mace}. In particular, \name successfully defends attacks by
UnlearnDiff, significantly outperforming the existing approaches, verifying its robustness.

% \subsection{}

\section{Conclusion}

In this work, we show that only fine-tuning CA layers for concept erasing in diffusion models could sometimes fail in preserving remaining concepts. As one solution, we proposed our framework, \name (Interpret-then-deactivate), simple and effective approach to remove the regulation constraint constraint  aiming to erase
target concepts while maintaining diverse remaining concepts. 
We integrate the Sparse Autoencoder (SAE) into the the diffusion models, making it capable of adaptively adjusting the text embeddings. To robustly erase target concepts without forgetting on remaining concepts,
we also comprehensively compare the unique features unique to the target concepts, and deactive them, removing the effect on the remaining concepts. Through
extensive experiments on erasure of celebrities, artistic styles, and explicit concepts, the empirical results ensure
the robust deletion of target concepts and protection of diverse remaining concepts.

\newpage 

\section{Impact Statements:}
% This paper presents work whose goal is to advance the field of Machine Learning. There are many potential societal consequences of our work, none of of which we feel must be specifically highlighted here. 
This paper presents work whose goal is to advance the field of Machine Learning. There are many potential societal consequences of our work, for example, improving the safety and ethical use of generative models by preventing the creation of harmful or explicit content. This could help mitigate the spread of inappropriate or harmful imagery, ensuring that AI systems align more closely with societal values and legal standards, particularly in areas such as content moderation, education, and creative industries.

\nocite{langley00}

\bibliography{main}
\bibliographystyle{icml2025}

\newpage
\appendix
\onecolumn
\section{Preliminaries on baseline approaches}
\label{sec:preliminary}
We consider six baseline approaches, including 4 fine-tuning-based approaches:ESD~\cite{gandikota2023erasing}, AC~\cite{kumari2023ablating}, AdvUn~\cite{zhang2024defensive}, Receler~\cite{huang2023receler}, and two inference-based approaches MACE~\cite{lu2024mace}, CPE~\cite{conceptpinpoint2025}.

\subsection{Cross-Attention in T2I Diffusion Model}
We first introduce the fundamentals of the cross-attention (CA) layer in T2I diffusion models, which is widely utilized in previous works.  
The CA layer serves as a key mechanism for integrating textual information, represented by text embeddings \(\mathbf{E}\), with image features, represented as image latents \( x_t \), within T2I diffusion models.
Specifically, the image feature $\phi(x_t)$ is linearly transformed into query matrix $Q=l_Q(\phi(x_t)$, while the text embedding is mapped into key matrix $K=l_K(\mathbf{E})$ and value matrix $V = l_V(\mathbf{E})$ through separate linear transformations. The attention map can be calculated as:
$$
M=\operatorname{Softmax}\left(\frac{Q K^T}{\sqrt{d}}\right) \in \mathbb{R}^{u \times(h \times w) \times n},
$$
where $d$ is the projection dimension, $u$ is the number of attention heads, $h\times w$ represents the spatial dimension of the image, and $n$ is the number of text tokens. The final output of the cross-attention is $MV$, representing the weighted average of the values in $V$.

% The cross attention value represents the weights of each token to each pixel of images, with a large weight representing the token is dominant to the pixel compared with other tokens. As a result, a higher attention map value indicates stronger relevance between the image region and the corresponding token. 

\subsection{Fine-tuning-based approaches}
% \begin{packeditemize}
\paragraph{ESD} is inspired by energy-based composition~\cite{du2020compositional,du2021unsupervised}. It aims to reduce the probability of generating images belonging to the target concept. Specifically, it aims to minimize
\begin{equation}
\label{equ:esd_prob}
    P_\theta(x) \propto \frac{P_{\thetao}(x)}{P_{\thetao}(c \mid x)^\eta},
\end{equation}
where $P_{\thetao}(x)$ represents the distribution generated by the original model, $c$ is the target concept to erase, and $\eta$ is scale parameter. 
Inspired by Equ.~\ref{equ:esd_prob}, and Tweedie’s formula~\cite{efron2011tweedie}
% the diffusion process of ESD is modified to 
\begin{equation}
\label{equ:esd_epsilon}
\tilde{\epsilon}_\theta\left(x_t \mid c_{e}\right) \leftarrow \epsilon_{\theta_o}\left(x_t \mid \emptyset\right)-\eta\left(\epsilon_{\theta_o}\left(x_t \mid c\right)-\epsilon_{\theta_o}\left(x_t \mid \emptyset\right)\right),
\end{equation}
The training loss to optimize $\theta$ is:
\begin{equation}
    \min _\theta \ell_{\operatorname{ESD}}(\theta, c):=\mathbb{E}\left[\left\|\epsilon_\theta\left(x_t \mid c\right)-\left(\epsilon_{\theta_0}\left(x_t \mid \emptyset\right)-\eta\left(\epsilon_{\theta_0}\left(x_t \mid c\right)-\epsilon_{\theta_0}\left(x_t \mid \emptyset\right)\right)\right)\right\|_2^2\right] .
\end{equation}

Empirical results show that updating only the CA layers (ESD-x) effectively erases target concepts while preserving the overall performance of remaining concepts. In contrast, fine-tuning all parameters (ESD-u) can lead to significant degradation of remaining concepts. In our implementation of ESD, \uline{we mainly implement ESD-x}, as it demonstrates good concept erasure while minimizing its impact on remaining concepts.

\paragraph{AdvUn} is designed based on ESD. It proposes a bi-level optimization approach that is robust to adversarial attacks. The upper-level optimization aims to minimize the unlearning objective, while lower-level optimization aims to find the optimized adversarial prompt $c^*$:
\begin{equation}
\begin{array}{lll}\underset{\boldsymbol{\theta}}{\operatorname{minimize}} & \ell_{\mathrm{u}}\left(\boldsymbol{\theta}, c^*\right) & \text { [Upper-level] } \\ \text { subject to } & c^*=\arg\min_{\left\|c^*-c\right\|_0 \leq \rho} \mathbb{E}\left[\left\|\epsilon_\theta\left(\mathbf{x}_t \mid c^*\right)-\epsilon_{\theta^*}\left(\mathbf{x}_t \mid c\right)\right\|_2^2\right] . & \text { [Lower-level] }
\end{array}
\end{equation}
To maintain the utility on remaining concepts, it adopts a regularization term to penalize the discrepancy between the original model and optimized model on a retain set $C_\text{retain}$. As a result, the upper-level optimization objective is:
$$
\ell_{\mathrm{u}}\left(\boldsymbol{\theta}, c^*\right)=\ell_{\mathrm{ESD}}\left(\boldsymbol{\theta}, c^*\right)+\gamma \mathbb{E}_{\tilde{c} \sim \mathcal{C}_{\text {retain }}}\left[\left\|\epsilon_{\boldsymbol{\theta}}\left(\mathbf{x}_t \mid \tilde{c}\right)-\epsilon_{\boldsymbol{\theta}_{\mathrm{o}}}\left(\mathbf{x}_t \mid \tilde{c}\right)\right\|_2^2\right]
$$

AdvUn tried to perform erasing on different layers within the text encoder and U-net in SD v1-4~\cite{rombach2022high}. Empirical results show that erasing within the text encoder has the best robustness against adversarial attacks. In our implementation of AdvUn, \uline{we perform erasing on the text encoder.}

\paragraph{Receler} is also designed based on ESD and adopts adversarial prompt learning to ensure erasure robustness. To reduce the impact on remaining concepts, it adopts a concept-localized regularization for erasing locality:
\begin{equation}
    \ell_{Reg}=\frac{1}{L} \sum_{l=1}^L\left\|o^l \odot(1-M)\right\|^2
\end{equation}
where $L$ is the number of U-Net’s layers, $\odot$ is the element-wise product, $o^l$ is the output of the eraser in the l-th layer, and $M$ is the mask of target concept in image and generated using GroundingDINO~\cite{liu2024grounding}

\paragraph{AC} is another approach for unlearning. It prevents the model from generating unwanted images by mapping target concepts to an anchor concept, which can be either a generic concept, such as 'dog' to replace 'English springer' or a null concept, such as an empty text prompt. Moreover, it also utilizes a regularization term to penalize the discrepancy between the original model and the optimized model on a set of retained concepts.
The training objective can be formulated as follows:
\begin{equation}
    \underset{\theta}{\operatorname{min}}\ell_{\mathrm{AC}}\left(\theta, c\right):=
    \mathbb{E}\left[
    \left\|\epsilon_{\theta}\left(x_t \mid c\right) -\epsilon_{\theta_o}\left(x_t \mid c_a\right) \right\|_2^2\right] + \gamma \mathbb{E}_{\tilde{c} \sim \mathcal{C}_{\text {retain }}}\left[\left\|\epsilon_{\boldsymbol{\theta}}\left(\mathbf{x}_t \mid \tilde{c}\right)-\epsilon_{\boldsymbol{\theta}_{\mathrm{o}}}\left(\mathbf{x}_t \mid \tilde{c}\right)\right\|_2^2\right],
\end{equation}
where $c_a$ is the surrogate concept.

% \end{packeditemize}

\subsection{Inference-based approaches}
\paragraph{MACE} adopts a closed-form solution to refine CA layers within U-net to erase unwanted knowledge. Briefly, it finds the linear projections $W'$ of \textit{keys} and \textit{values} in the cross-attention layers such that:
\begin{equation}
    \mathbf{W}_{\text {new }}=\arg \min _{\mathbf{W}^{\prime}} \sum_{n=1}^N\left\|\mathbf{W}^{\prime} \mathbf{E}_{\text {tar }}^n-\mathbf{W}_{\text {o}} \mathbf{E}_{\text {sur }}^n\right\|_F^2+\lambda \sum_{m=1}^M\left\|\mathbf{W}^{\prime} \mathbf{E}_{\text {retain }}^m-\mathbf{W}_{\text {o}} \mathbf{E}_{\text {retain }}^m\right\|_F^2,
\end{equation}
where $\mathbf{W}_{\text{o}}$ is the original key/value projection. $\mathbf{E}_{\text {tar}}, \mathbf{E}_{\text{sur}}$ and $\mathbf{E}_{\text {retain}}$ represent the text embedding of target, surrogate, and retain concepts, respectively. To mitigate the impact on the overall parameters, MACE inserts LoRA modules into the CA layers of the model for each target concept. Then, the multiple LoRAs are integrated by a loss function to enable erasing multiple concepts.

\paragraph{CPE} also works on the linear projections of \textit{keys} and \textit{values} in the cross-attention layers. It inserts a customized modular, named residual attention gate (ResAG), into each CA layer within U-net. ResAG is trained to make the projection output of $\mathbf{E}_{\mathrm{tar}}$ similar to the output of $\mathbf{E}_{\mathrm{sur}}$. 
Formally, the erasing objective is:
\begin{equation}
    \underset{R_{\mathrm{tar}}}{\operatorname{min}}\ell_{\text{era}}=\mathbb{E}_{\left(\mathbf{E}_{\mathrm{tar}}, \mathbf{E}_{\mathrm{sur}}\right)}\left\|\left(\mathbf{W}\mathbf{E}_{\mathrm{tar}}+R_{\mathrm{tar}}\left(\mathbf{E}_{\mathrm{tar}}\right)\right)-\left(\mathbf{W} \mathbf{E}_{\mathrm{sur}}-\eta \mathbf{W}\left(\mathbf{E}_{\mathrm{tar}}-\mathbf{E}_{\mathrm{sur}}\right)\right)\right\|^2,
\end{equation}
where $R_{\mathrm{tar}}$ is the ResAG for target concept. To prevent undesirable degradation on remaining concepts, CPE adopts a regularization term to minimize the deviation induced by ResAG on retain concepts:
\begin{equation}
    \mathcal{L}_{\text {att}}=\mathbb{E}_{\mathbf{E}_{\text {retain}}}\left\|R_{\mathrm{tar}}\left(\mathbf{E}_{\text {retain}}\right)\right\|_F.
\end{equation}
Each ResAG is trained specifically to one target concept. Therefore, to erase multiple concepts, it is required to train multiple ResAGs.

\section{Training details of SAE}
\label{sec:sae_training}

% \begin{figure}[ht]
% \centering
% \begin{tabular}{ccc}
%     \subfloat{
%         \includegraphics[width=0.45\textwidth,valign=c]{./plots/feature_density_celeb.png}
%     } 
%     \subfloat{
%         \includegraphics[width=0.45\textwidth,valign=c]{./plots/feature_density_nudity.png}
%     } 
% \end{tabular}
% \caption{The log feature density is calculated as \(\log(n/N + 10^{-9}) \), where \( n \) represents the number of tokens that activate the feature, and \( N \) denotes the total number of tested tokens.}
% \label{fig:feature_density}
% \end{figure}

We train an SAE using the output of the 8th transformer block of the text encoder \texttt{layers.8}, which we experimentally show has the best performance. We set $K=64$ and $d_{\text{hid}}=2^{19}$ following~\cite{gao2024scalingevaluatingsparseautoencoders}. We adopt Adam~\cite{kinga2015method} as the optimizer with the learning rate of $5e-5$ and a constant scheduler without warmup.

Following~\citet{gao2024scalingevaluatingsparseautoencoders}, we set $\alpha=\frac{1}{32}$ and $K_{\text{aux}}=256$. We train the SAE while simultaneously generating training samples with the text encoder, which does not require additional storage space to save samples. The batch size of prompts input into the text encoder is 50, which results in about 1000 samples to train SAE each time. 

We train SAE on a single H100. For celebrity and artistic style erasure, where we train an SAE using celebrity and artist styles and the captions of COCO-30K, the training time is 56 minutes. For nudity erasure, we train an SAE using the captions of COCO-30 and 10K prompts from DiffusionDB~\cite{DiffusionDB} with an NSFW score larger than 0.8, the training time is 40 minutes.

\section{Efficiency study}
The structure of SAE used in \name is very simple, with only two linear layers, two bias layers, and the $\operatorname{TopK}$ activation function. 
While it contains a large number of parameters due to the large hidden size, it is efficient during inference as it mainly requires two matrix multiplication operations.

We report the inference time of SAE along with its time ratio relative to the entire image generation process in Table~\ref{tab:efficiency}. The inference time is measured per prompt, which consists of 77 tokens (the maximum length allowed in SD v1.4 and SD v2.1). The time required to generate a single image is computed by repeating the process 10 times and taking the average. The number of inference steps is set to 50.
The results indicate that SAE is highly efficient, accounting for less than 1\% of the total image generation time. 

We note that unlike other module-based approaches that require increased inference time as more concepts are erased~\cite{lyu2024one,conceptpinpoint2025}, the inference time of \name is independent of the number of concepts being erased.

\begin{table}[ht]
    \centering
    \begin{tabular}{c|c|c}
    \toprule
         & SD v1.4 & SD v2.1\\\midrule
        % \# of Params  800M&  & \\
       % Param Ratio  &  & \\
       Inference Time (1000 prompts)  & 5.05s & 6.37s \\
       Time Ratio (per prompt)  & 0.22\%  & 0.13\% \\ \bottomrule
       
    \end{tabular}
    \caption{\textbf{Efficiency study}. The inference time of SAE along with its time ratio relative to the entire image generation process.}
    \label{tab:efficiency}
\end{table}

\section{Implementation details}

\subsection{Celebrity Erasure}

We select 50 celebrities from 200 celebrities provide in MACE~\cite{lu2024mace} as target concepts to erase.
The celebrities can be accurately generated by Stable Diffusion v1.4~\cite{rombach2022high}, which have over 99\% accuracy of the GIPHY Celebrity Detector (GCD)~\cite{celebritydetector}. The 50 target celebrities are listed in Table~\ref{tab:celebrity_erase}.
We also select 100 celebrities as remaining concepts to preserve, as listed in Table~\ref{tab:celebrity_preserve}.
To generate their images, we used 5 prompt templates with 5 random seeds (1-5). The prompt templates are distinct for celebrities and artistic styles. We used 0 as a seed generating 5 images from a prompt for characters. The prompt templates are listed in Table~\ref{tab:template}.

To select features specific to the target celebrities, we adopt the remaining 100 celebrities as well as 1000 captions from COCO-30K as the retain set. 

In the case of celebrities erasure, we set the following negative prompts to improve image quality: \textit{"bad anatomy, watermark, extra digit, signature, worst quality, jpeg artifacts, normal
quality, low quality, long neck, lowres, error, blurry, missing fingers, fewer digits, missing arms, text,
cropped, humpbacked, bad hands, username"}

% Similarly, Table \ref{tab:celeb settings} presents different settings for concept erasure when targeting a specific celebrity. Each method employs a different surrogate concept and defines remaining concepts to ensure that the model retains meaningful knowledge. To ensure MACE's~\cite{lu2024mace} ability to retain concepts, we use its publicly available pre-trained model for unlearning 10 celebrities.

% \begin{table*}[h]
%     \centering
%     \caption{{List of concept settings for erasing target celebrities.}}
%     \label{tab:celeb settings}
%     % \renewcommand{\arraystretch}{1.3}
%     \resizebox{\textwidth}{!}{
%     \begin{tabular}{|l|l|l|}
%         \hline
%         \textbf{Method} & \textbf{Surrogate Concept} & \textbf{Remain Concepts} \\
%         \hline
%         ESD     & -      & -  \\
%         AC      & ''person'' & 100 celebrities \\
%         AdvUn   & '' ''      & 100 celebrities  \\
%         Receler & '' ''      & 100 celebrities  \\
%         MACE    & ''art''    & 100 celebrities + all captions from MS-COCO \\
%         CPE     & ''a person'' & 50 similar celebrities from 500 celebrities + 100 similar words with ‘celebrities’ \\
%         \hline
%     \end{tabular}
%     }
% \end{table*}

\begin{table}[!ht]
\centering
\caption{List of target celebrities. We adopt the same 50 celebrities following~\cite{conceptpinpoint2025}. The selected celebrities have over 99\% accuracy by the GIPHY Celebrity
Detector (GCD)~\cite{celebritydetector}.}
\label{tab:celebrity_erase}
\begin{tabular}{|c|p{0.7\textwidth}|}
\toprule
\# of Celebrities to be erased & Celebrity \\
\midrule
50 & 'Adam Driver', 'Adriana Lima', 'Amber Heard', 'Amy Adams', 'Andrew Garfield', 'Angelina Jolie', 'Anjelica Huston', 'Anna Faris', 'Anna Kendrick', 'Anne Hathaway', 'Arnold Schwarzenegger', 'Barack Obama', 'Beth Behrs', 'Bill Clinton', 'Bob Dylan', 'Bob Marley', 'Bradley Cooper', 'Bruce Willis', 'Bryan Cranston', 'Cameron Diaz', 'Channing Tatum', 'Charlie Sheen', 'Charlize Theron', 'Chris Evans', 'Chris Hemsworth','Chris Pine', 'Chuck Norris', 'Courteney Cox', 'Demi Lovato', 'Drake', 'Drew Barrymore', 'Dwayne Johnson', 'Ed Sheeran', 'Elon Musk', 'Elvis Presley', 'Emma Stone', 'Frida Kahlo', 'George Clooney', 'Glenn Close', 'Gwyneth Paltrow', 'Harrison Ford', 'Hillary Clinton', 'Hugh Jackman', 'Idris Elba', 'Jake Gyllenhaal', 'James Franco', 'Jared Leto', 'Jason Momoa', 'Jennifer Aniston', 'Jennifer Lawrence'\\
\bottomrule 
\end{tabular}
\end{table}

\begin{table}[!ht]
\centering
\caption{List of celebrities to preserve. We adopt the same 100 celebrities following~\cite{conceptpinpoint2025}. The selected celebrities have over 99\% accuracy by the GIPHY Celebrity
Detector (GCD)~\cite{celebritydetector}.}
\label{tab:celebrity_preserve}
\begin{tabular}{|c|p{0.7\textwidth}|}
\toprule
\# of Celebrities to be preserve & Celebrity \\
\midrule
100 & 'Aaron Paul', 'Alec Baldwin', 'Amanda Seyfried', 'Amy Poehler', 'Amy Schumer', 'Amy Winehouse', 'Andy Samberg', 'Aretha Franklin', 'Avril Lavigne', 'Aziz Ansari', 'Barry Manilow', 'Ben Affleck', 'Ben Stiller', 'Benicio Del Toro', 'Bette Midler', 'Betty White', 'Bill Murray', 'Bill Nye', 'Britney Spears', 'Brittany Snow', 'Bruce Lee', 'Burt Reynolds', 'Charles Manson', 'Christie Brinkley', 'Christina Hendricks', 'Clint Eastwood', 'Countess Vaughn', 'Dane Dehaan', 'Dakota Johnson', 'David Bowie', 'David Tennant', 'Denise Richards', 'Doris Day', 'Dr Dre', 'Elizabeth Taylor', 'Emma Roberts', 'Fred Rogers', 'George Bush', 'Gal Gadot', 'George Takei', 'Gillian Anderson', 'Gordon Ramsey', 'Halle Berry', 'Harry Dean Stanton', 'Harry Styles', 'Hayley Atwell', 'Heath Ledger', 'Henry Cavill', 'Jackie Chan', 'Jada Pinkett Smith', 'James Garner', 'Jason Statham', 'Jeff Bridges', 'Jennifer Connelly', 'Jensen Ackles', 'Jim Morrison', 'Jimmy Carter', 'Joan Rivers', 'John Lennon', 'Jon Hamm', 'Judy Garland', 'Julianne Moore', 'Justin Bieber', 'Kaley Cuoco', 'Kate Upton', 'Keanu Reeves', 'Kim Jong Un', 'Kirsten Dunst', 'Kristen Stewart', 'Krysten Ritter', 'Lana Del Rey', 'Leslie Jones', 'Lily Collins', 'Lindsay Lohan', 'Liv Tyler', 'Lizzy Caplan', 'Maggie Gyllenhaal', 'Matt Damon', 'Matt Smith', 'Matthew Mcconaughey', 'Maya Angelou', 'Megan Fox', 'Mel Gibson', 'Melanie Griffith', 'Michael Cera', 'Michael Ealy', 'Natalie Portman', 'Neil Degrasse Tyson', 'Niall Horan', 'Patrick Stewart', 'Paul Rudd', 'Paul Wesley', 'Pierce Brosnan', 'Prince', 'Queen Elizabeth', 'Rachel Dratch', 'Rachel Mcadams', 'Reba Mcentire', 'Robert De Niro'\\
\bottomrule 
\end{tabular}
\end{table}

\subsection{Artist Style Erasure}
We select 100 artist styles as target concepts to erase, and 100 artist styles as remaining concepts to preserve. The 100 target artistic styles are listed in Table~\ref{tab:style_erase} and the remaining concepts are listed in Table~\ref{tab:style_preserve}. To generate their images, we used 5 prompt templates with 5 random seeds (1-5). The prompt templates are listed in Table~\ref{tab:template}, which are different from celebrity erasure.

To select features specific to the target artist styles, we adopt the remaining 100 artist styles as well as 1000 captions from COCO-30K as the retain set. 

% \begin{table*}[h]
%     \centering
%     \caption{{List of concept settings for erasing target artistic styles.}}
%     \label{tab:style settings}
%     % \renewcommand{\arraystretch}{1.3}
%     \resizebox{\textwidth}{!}{
%     \begin{tabular}{|l|l|l|}
%         \hline
%         \textbf{Method} & \textbf{Surrogate Concept} & \textbf{Retain Concepts} \\
%         \hline
%         ESD     & - & -  \\
%         AC      & '' '' & 100 styles  \\
%         AdvUn   & '' '' & 100 styles  \\
%         Receler & '' '' & 100 styles  \\
%         MACE    & ''art'' & 100 styles + all captions from MS-COCO  \\
%         CPE     & ''real photograph'' & 50 similar styles from 1734 styles + 100 similar words with 'famous artists'  \\
%         \hline
%     \end{tabular}
%     }
% \end{table*}

\begin{table}[ht]
\centering
\caption{List of target artist styles. We adopt the same 100 artist styles following~\cite{conceptpinpoint2025}. All artistic styles in these images were successfully generated using SD v1.4.}
\label{tab:style_erase}
\begin{tabular}{|c|p{0.7\textwidth}|}
\toprule
\# of Artist Styles to be erased & Artist Style \\
\midrule
100 & 'Brent Heighton', 'Brett Weston', 'Brett Whiteley', 'Brian Bolland', 'Brian Despain', 'Brian Froud', 'Brian K. Vaughan', 'Brian Kesinger', 'Brian Mashburn', 'Brian Oldham', 'Brian Stelfreeze', 'Brian Sum', 'Briana Mora', 'Brice Marden', 'Bridget Bate Tichenor', 'Briton Riviere', 'Brooke Didonato', 'Brooke Shaden', 'Brothers Grimm', 'Brothers Hildebrandt', 'Bruce Munro', 'Bruce Nauman', 'Bruce Pennington', 'Bruce Timm', 'Bruno Catalano', 'Bruno Munari', 'Bruno Walpoth', 'Bryan Hitch', 'Butcher Billy', 'C. R. W. Nevinson', 'Cagnaccio Di San Pietro', 'Camille Corot', 'Camille Pissarro', 'Camille Walala', 'Canaletto', 'Candido Portinari', 'Carel Willink', 'Carl Barks', 'Carl Gustav Carus', 'Carl Holsoe', 'Carl Larsson', 'Carl Spitzweg', 'Carlo Crivelli', 'Carlos Schwabe', 'Carmen Saldana', 'Carne Griffiths', 'Casey Weldon', 'Caspar David Friedrich', 'Cassius Marcellus Coolidge', 'Catrin WelzStein', 'Cedric Peyravernay', 'Chad Knight', 'Chantal Joffe', 'Charles Addams', 'Charles Angrand', 'Charles Blackman', 'Charles Camoin', 'Charles Dana Gibson', 'Charles E. Burchfield', 'Charles Gwathmey', 'Charles Le Brun', 'Charles Liu', 'Charles Schridde', 'Charles Schulz', 'Charles Spencelayh', 'Charles Vess', 'Charles-Francois Daubigny', 'Charlie Bowater', 'Charline Von Heyl', 'Cha "im Soutine', 'Chen Zhen', 'Chesley Bonestell', 'Chiharu Shiota', 'Ching Yeh', 'Chip Zdarsky', 'Chris Claremont', 'Chris Cunningham', 'Chris Foss', 'Chris Leib', 'Chris Moore', 'Chris Ofili', 'Chris Saunders', 'Chris Turnham', 'Chris Uminga', 'Chris Van Allsburg', 'Chris Ware', 'Christian Dimitrov', 'Christian Grajewski', 'Christophe Vacher', 'Christopher Balaskas', 'Christopher Jin Baron', 'Chuck Close', 'Cicely Mary Barker', 'Cindy Sherman', 'Clara Miller Burd', 'Clara Peeters', ‘Clarence Holbrook Carter', ‘Claude Cahun', 'Claude Monet', ‘Clemens Ascher'\\
\bottomrule 
\end{tabular}
\end{table}

\begin{table}[!h]
\centering
\caption{Prompt templates used for celebrity image generation. For each prompt, we generate 5 images with seed ranges from 1-5. Additionally, for celebrities, we set the following negative prompts to improve image quality: \textit{"bad anatomy, watermark, extra digit, signature, worst quality, jpeg artifacts, normal
quality, low quality, long neck, lowres, error, blurry, missing fingers, fewer digits, missing arms, text,
cropped, humpbacked, bad hands, username"}} 
\label{tab:template}
\begin{tabular}{|c|c|p{0.75\textwidth}|} % Fixed vertical bar issue
\toprule
Domain & Seed & Prompt \\
\midrule
Celebrity & 1-5 & 
\textit{A portrait of \{celebrity name\}} \newline
\textit{A sketch of \{celebrity name\}} \newline
\textit{An oil painting of \{celebrity name\}} \newline
\textit{\{celebrity name\} in an official photo} \newline
\textit{An image capturing \{celebrity name\} at a public event} \\
\midrule
Artist Style & 1-5 &
\textit{Image in the style of \{artist name\}} \newline
\textit{Art inspired by \{artist name\}} \newline
\textit{Painting in the style of $\{$ artist name $\}$} \newline
\textit{A reproduction of art by \{artist name $\}$} \newline
\textit{A famous artwork by \{artist name\}} \\
\bottomrule 
\end{tabular}
\end{table}

\begin{table}[ht]
\centering
\caption{List of artist styles to preserve. We adopt the same 100 artist styles following~\cite{conceptpinpoint2025}. All artistic styles in these images were successfully generated using SD v1.4.}
\label{tab:style_preserve}
\begin{tabular}{|c|p{0.7\textwidth}|}
\toprule
\# of Artist styles to preserve & Artist Style  \\
\midrule
100 & ‘A.J.Casson', ‘Aaron Douglas', ‘Aaron Horkey', ‘Aaron Jasinski', ‘Aaron Siskind', ‘Abbott Fuller Graves', 'Abbott Handerson Thayer', 'Abdel Hadi Al Gazzar', 'Abed Abdi', 'Abigail Larson', 'Abraham Mintchine', 'Abraham Pether', 'Abram Efimovich Arkhipov', 'Adam Elsheimer', 'Adam Hughes', 'Adam Martinakis', 'Adam Paquette', 'Adi Granov', 'Adolf Hiremy-Hirschl', 'Adolph Got- 'tlieb', 'Adolph Menzel', 'Adonna Khare', 'Adriaen van Ostade', 'Adriaen van Outrecht', 'Adrian Donoghue', 'Adrian Ghenie', 'Adrian Paul Allinson', 'Adrian Smith', 'Adrian Tomine', 'Adrianus Eversen', 'Afarin Sajedi', 'Affandi', 'Aggi Erguna', 'Agnes Cecile', 'Agnes Lawrence Pelton', 'Agnes Martin', 'Agostino Arrivabene', 'Agostino Tassi', 'Ai Weiwei', 'Ai Yazawa', 'Akihiko Yoshida', 'Akira Toriyama', 'Akos Major', 'Akseli Gallen-Kallela', 'Al Capp', 'Al Feldstein', 'Al Williamson', 'Alain Laboile', 'Alan Bean', 'Alan Davis', 'Alan Kenny', 'Alan Lee', 'Alan Moore', 'Alan Parry', 'Alan Schaller', 'Alasdair McLellan', 'Alastair Magnaldo', 'Alayna Lemmer', 'Albert Benois', 'Albert Bierstadt', 'Albert Bloch', 'Albert Dubois-Pillet', 'Albert Eckhout', 'Albert Edelfelt', 'Albert Gleizes', 'Albert Goodwin', 'Albert Joseph Moore', 'Albert Koetsier', 'Albert Kotin', 'Albert Lynch', 'Albert Marquet', 'Albert Pinkham Ryder', 'Albert Robida', 'Albert Servaes', 'Albert Tucker', 'Albert Watson', 'Alberto Biasi', 'Alberto Burri', 'Alberto Giacometti', 'Alberto Magnelli', 'Alberto Seveso', 'Alberto Sughi', 'Alberto Vargas', 'Albrecht Anker', 'Albrecht Durer', 'Alec Soth', 'Alejandro Burdisio', 'Alejandro Jodorowsky', 'Aleksey Savrasov', 'Aleksi Briclot', 'Alena Aenami', 'Alessandro Allori', 'Alessandro Barbucci', 'Alessandro Gottardo', 'Alessio Albi', 'Alex Alemany', 'Alex Andreev' 'Alex Colville', 'Alex Figini', 'Alex Garant'\\
\bottomrule 
\end{tabular}
\end{table}

\section{Ablation studies for \name}
\subsection{The Effect of strength $\lambda$}

\begin{figure}[hbp]
\centering
\begin{tabular}{ccc}
    \subfloat{
        \includegraphics[width=0.45\textwidth,valign=c]{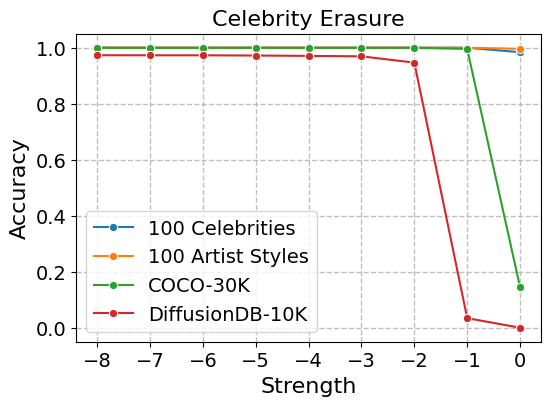}
    } 
    \subfloat{
        \includegraphics[width=0.45\textwidth,valign=c]{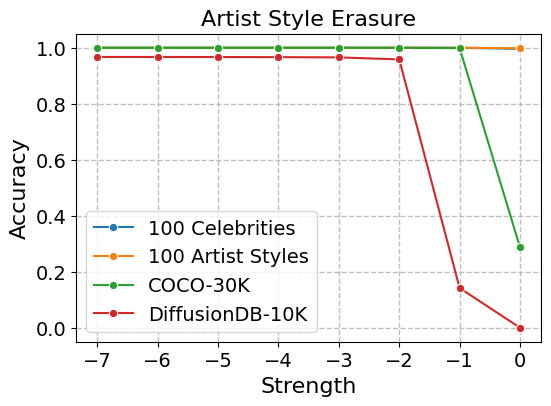}
    } 
\end{tabular}
\caption{\textbf{Ablation study on the effect of strength \( \lambda \) in SAE classification for distinguishing target and remaining concepts.} The threshold \( \tau \) is set to ensure zero false negatives; therefore, we primarily report accuracy on the remaining concepts. A higher accuracy indicates that SAE would not misclassify normal concepts as target concepts.}
\label{fig:strength}
\end{figure}

In this section, we investigate the impact of the strength parameter \( \lambda \) on SAE's effectiveness as a classifier for distinguishing between target and remaining concepts, as well as its ability to erase unwanted knowledge during text encoder inference.

We vary \( \lambda \) from -8 to 0 and conduct experiments on celebrity and artistic style erasure tasks. Specifically, we measure SAE’s accuracy in correctly identifying remaining concepts, conditioned on successfully identifying all target concepts. This is analogous to the true negative rate (TN) under the condition that the false negative rate (FN) is zero.
A higher value indicates greater effectiveness, ensuring that SAE would not misclassify normal concepts as target concepts.  
The results, presented in Figure~\ref{fig:strength}, demonstrate that our approach is robust to the choice of \( \tau \). When \( \tau < -1 \), SAE performs well across the remaining 100 celebrities, 100 artistic styles, and COCO-30K. Moreover, when \( \tau < -2 \), SAE correctly classifies over 95\% of prompts in DiffusionDB-10K as normal concepts.

\begin{figure}[hbp]
\centering
\begin{tabular}{c} 
    \subfloat{
        \includegraphics[width=0.81\textwidth,valign=c]{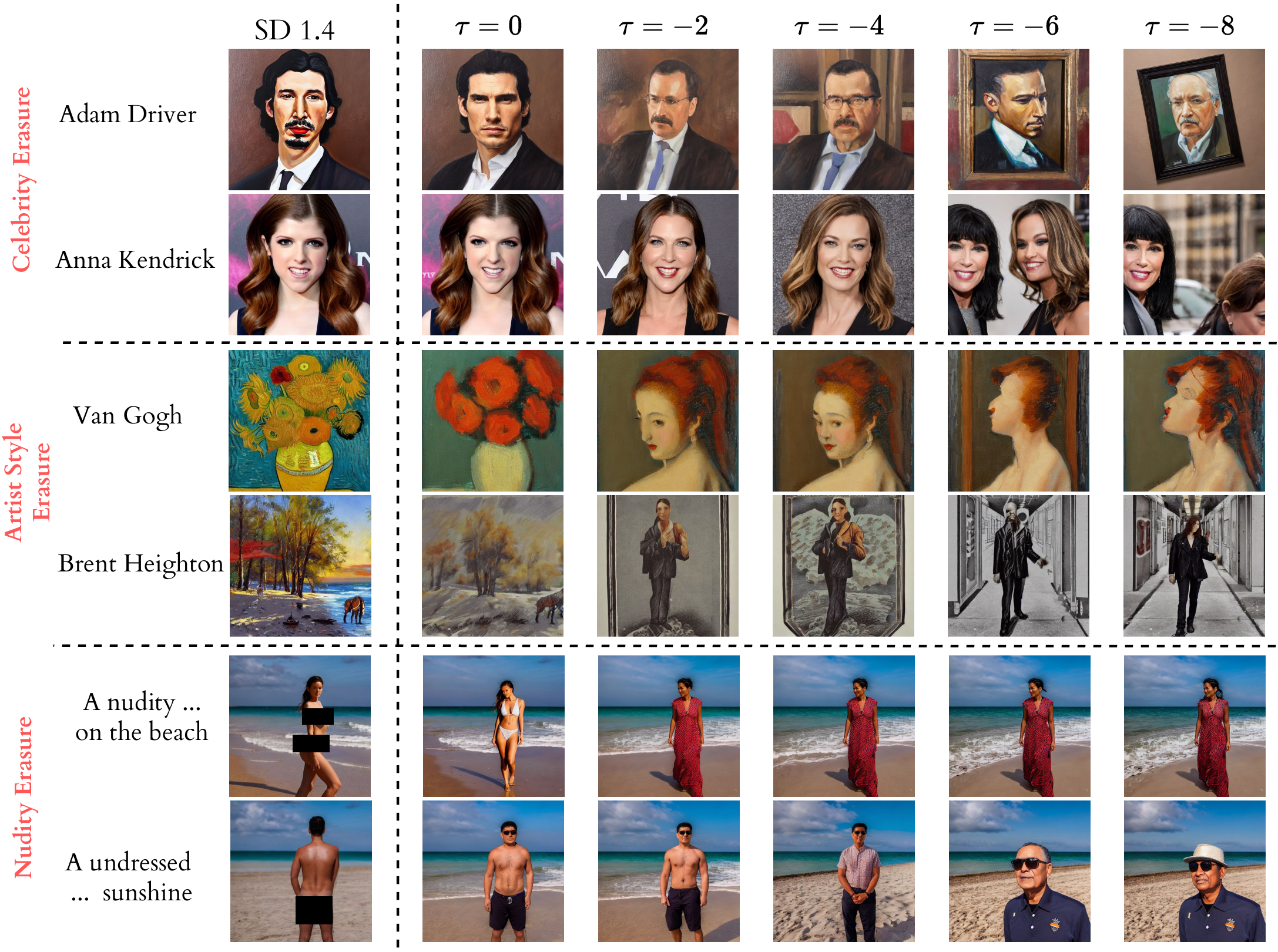}
    } 
\end{tabular}
\caption{\textbf{Ablation study on the effect of strength \( \lambda \) in erasing unwanted knowledge.} Experiments on SD v1.4~\cite{rombach2022high}.}
\label{fig:sd14}
\end{figure}

\begin{figure}[th]
\centering
\begin{tabular}{c} 
    \subfloat{
    \includegraphics[width=0.8\textwidth,valign=c]{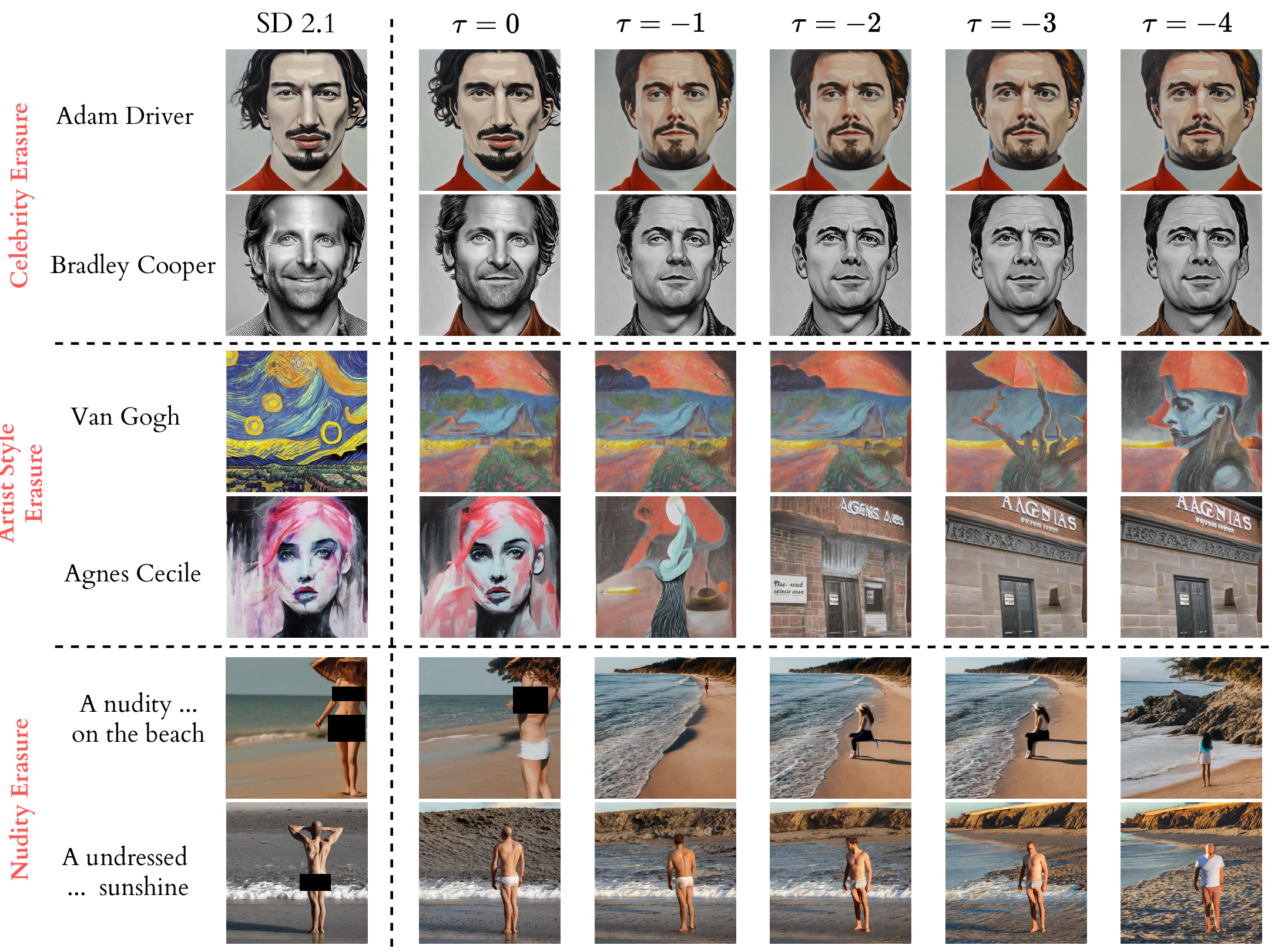}
    } 
\end{tabular}
\caption{\textbf{Ablation study on the effect of strength \( \lambda \) in erasing unwanted knowledge.} Experiments on SD v2.1~\cite{sd2}.}
\label{fig:sd21}
\end{figure}

To demonstrate the effectiveness of $\lambda$ in erasing target knowledge, we present generated images of target concepts with different $\lambda$. Figure~\ref{fig:sd14} show the results experiment on sd v1.4~\cite{rombach2022high} and and Figure~\ref{fig:sd21} show the results experiment on sd v2.1~\cite{sd2}. The results show that $\lambda=-2$ is sufficient to erase knowledge about celebrity and artist style concepts. The knowledge about "nudity" can be erased when $\lambda=-4$.

\subsection{Selection of Residual Stream to perform SAE.}
The text encoder consists of 12 transformer blocks (\texttt{layers.0-11}) connected sequentially. Each residual stream of these layers can be used to train SAE. However, since different layers may capture different types of knowledge~\cite{jin2024exploring}, applying SAE to different residual streams results in varying performance.  

To identify the most approximate layer for concept erasing, we conduct experiments on different residual streams. The training setup follows the details provided in Appendix~\ref{sec:sae_training}, with the only variation being the choice of residual stream. We evaluate the task of erasing 50 celebrities and 100 artist styles separately.
The remaining concepts consist of 100 other celebrities and 100 other artistic styles, COCO-30K, and DiffusionDB-10K. 
For efficiency, similar to Figure~\ref{fig:strength}, we use accuracy as a metric to assess the impact of different layers. This metric quantifies the proportion of correctly identified remaining concepts when SAE is used as a classifier.

\begin{figure}[htbp]
\centering
\begin{tabular}{ccc}
    \subfloat{
        \includegraphics[width=0.45\textwidth,valign=c]{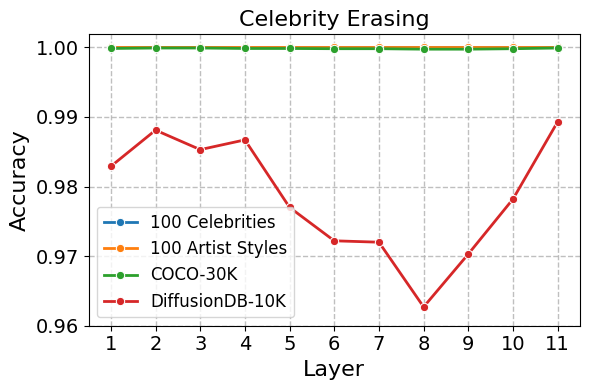}
    } 
    \subfloat{
        \includegraphics[width=0.45\textwidth,valign=c]{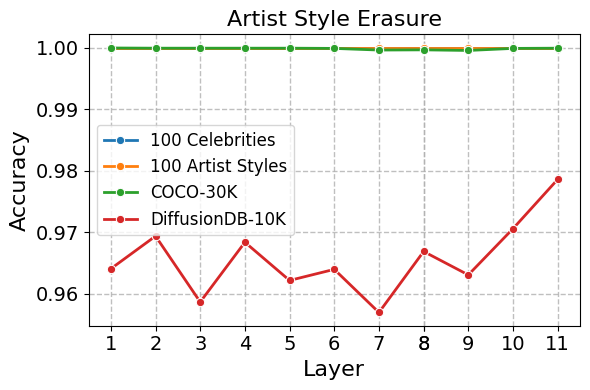}
    } 
\end{tabular}
\caption{The log feature density is calculated as \(\log(n/N + 10^{-9}) \), where \( n \) represents the number of tokens that activate the feature, and \( N \) denotes the total number of tested tokens.}
\label{fig:layers_acc}
\end{figure}

The results are summarized in Table~\ref{fig:layers_acc}. For remaining celebrities, artistic styles, and COCO-30K, the accuracy is approximately 100\%, demonstrating that SAE, as a classifier, effectively identifies remaining concepts regardless of the training layer. Even for DiffusionDB-10K, which exhibits the lowest performance, the accuracy remains at least 96\%.

\begin{figure}[h]
\centering
\begin{tabular}{c} 
    \subfloat{
        \includegraphics[width=0.85\textwidth,valign=c]{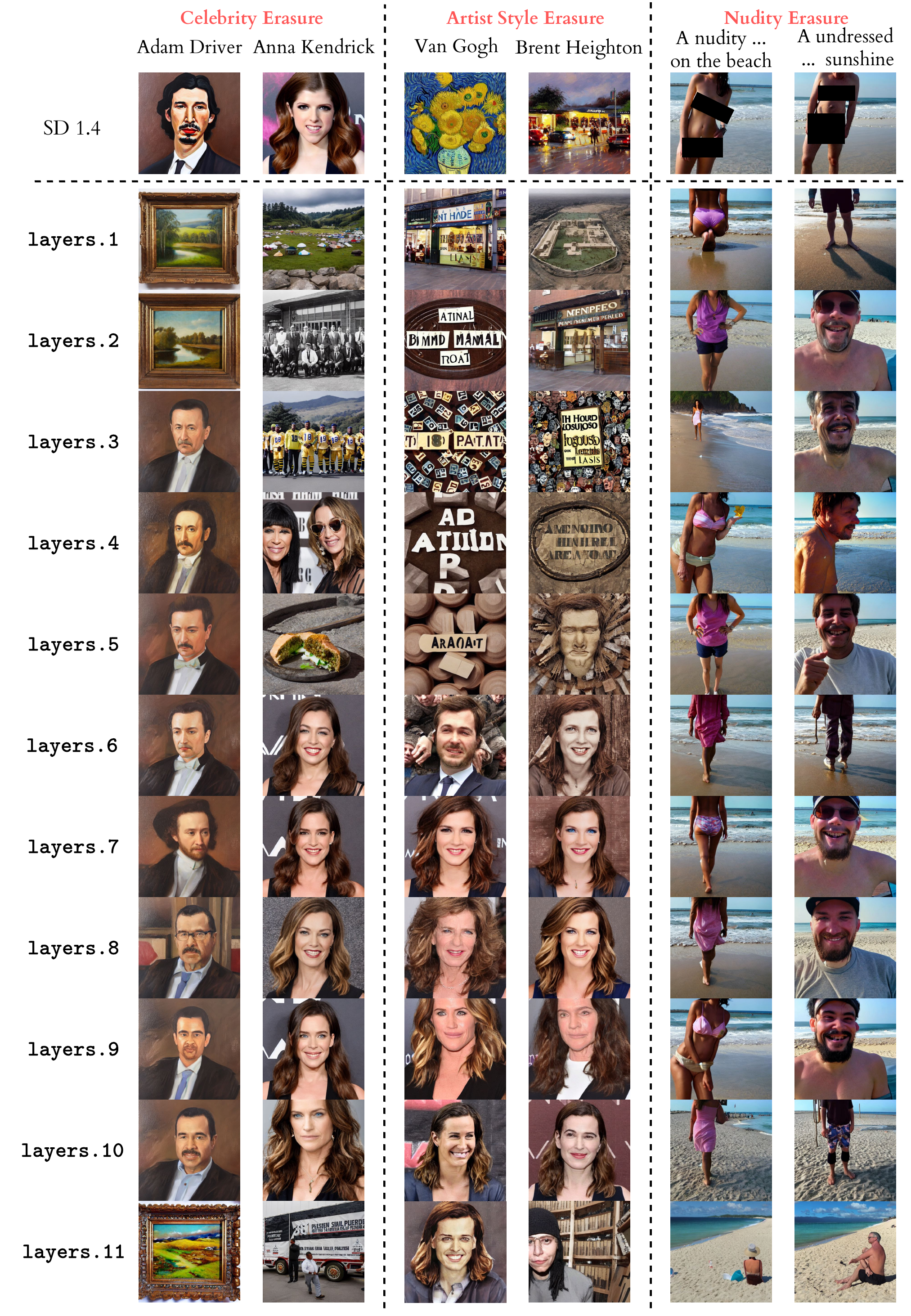}
    } 
\end{tabular}
\caption{Qualitative results of performing \name on different transformer blocks.}
\label{fig:sd14_layer}
\end{figure}

We also provide quantitative results to better understand the effects of erasing at different layers. Our experiments cover three domains: celebrities, artistic styles, and nudity content. The erasure strength \( \tau \) is set to -6 for celebrities and nudity and -2 for artistic styles.  
As shown in Figure~\ref{fig:sd14_layer}, all layers can be used to erase target concepts. However, certain layers (e.g., \texttt{layers.2-4}) still generate images with exposed chests when attempting to erase the concept of "nudity".

For the celebrity domain, we use the prompts \textit{"An oil painting of Adam Driver"} and \textit{"Anna Kendrick in an official photo."}. The results indicate that applying SAE to \texttt{layers.6-10} still allows the model to generate a person within the corresponding context (\textit{"oil painting"} and \textit{"official photo"}), suggesting that erasing at these layers preserves the structural integrity of the scene while removing the target identity. For artistic style erasure, applying SAE to \texttt{layers.6-10} also results in the generation of a photo of a person, even though the original images did not contain any people.

\begin{figure}[thbp]
\centering
\begin{tabular}{c}
    \subfloat[Results of SAE trained with \textbf{$d_{\text{hid}}=2^{15}$}]{
        \includegraphics[width=0.3\textwidth,valign=c]{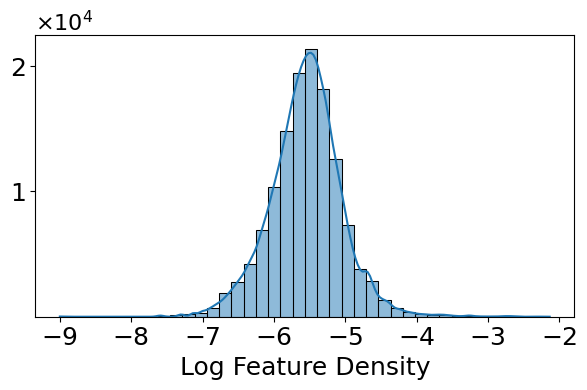}
        \includegraphics[width=0.3\textwidth,valign=c]{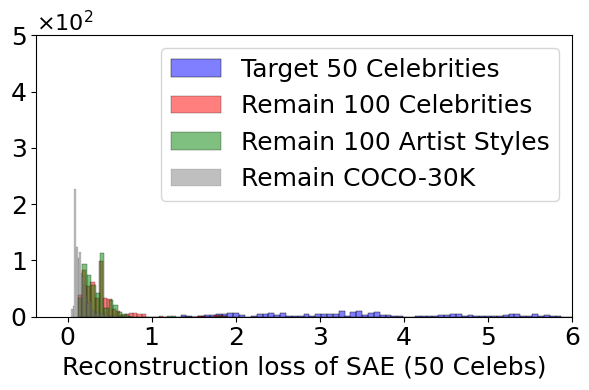}
        \includegraphics[width=0.3\textwidth,valign=c]{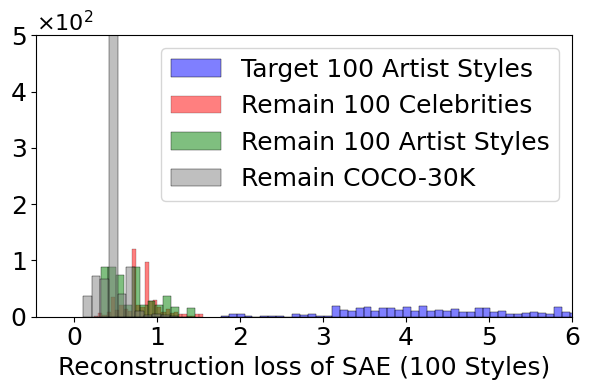}
    }\\
    \subfloat[Results of SAE trained with $d_{\text{hid}}=2^{17}$]{
        \includegraphics[width=0.3\textwidth,valign=c]{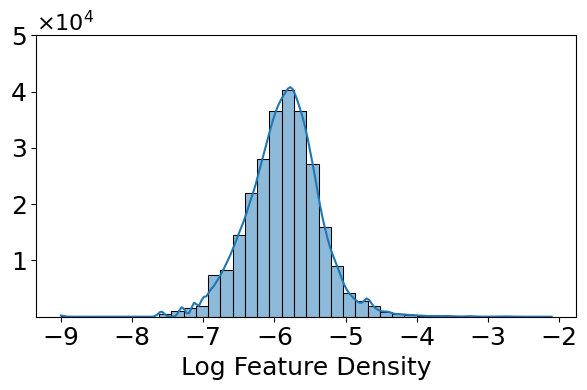}
        \includegraphics[width=0.3\textwidth,valign=c]{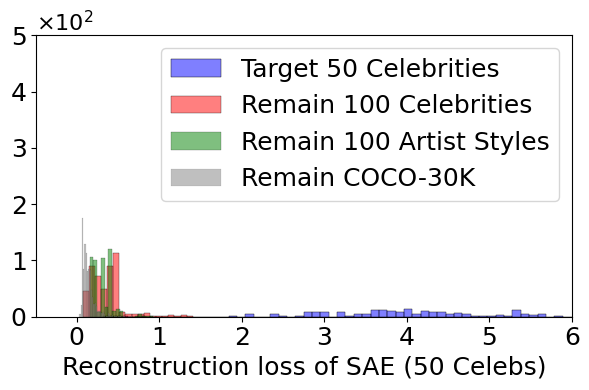}
        \includegraphics[width=0.3\textwidth,valign=c]{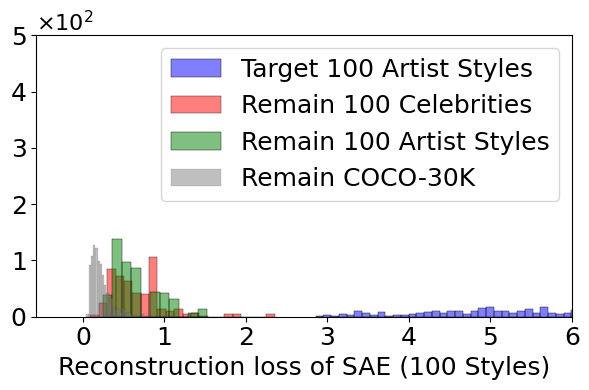}
    } \\
    \subfloat[Results of SAE trained with \textbf{$d_{\text{hid}}=2^{19}$}]{
        \includegraphics[width=0.3\textwidth,valign=c]{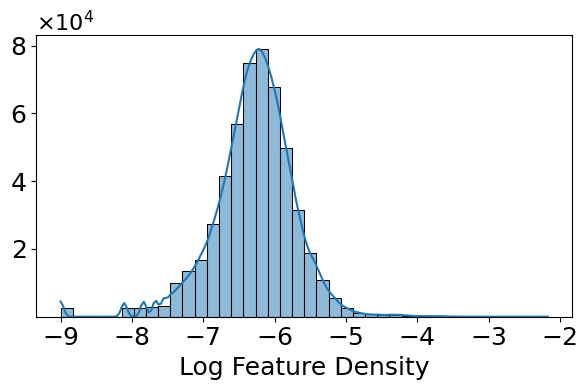}
        \includegraphics[width=0.3\textwidth,valign=c]{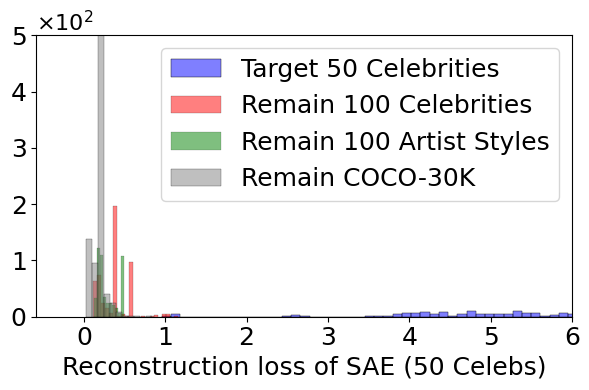}
        \includegraphics[width=0.3\textwidth,valign=c]{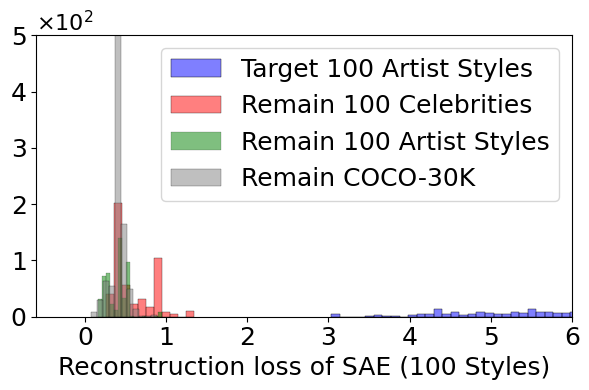}
         
    }  
\end{tabular}

\caption{The log feature density and the reconstruction loss for SAEs trained with different hidden sizes. }
\label{fig:hidden_size}
\end{figure}

\subsection{The effect of hidden size $d_{\text{hid}}$}

We investigate the effect of SAE with different hidden sizes for concept erasing. We vary  $d_{\text{hid}}$ from $2^{15}$ to $2^{19}$, which is about $43$ to $683$ times larger than the hidden dimension of SD v1-4~\cite{rombach2022high}. We present the feature density of the SAE as well as the reconstruction loss for target and remaining concepts in Figure~\ref{fig:hidden_size}. 
The results indicate that the log feature density is relatively higher for smaller hidden sizes compared to larger hidden sizes, suggesting that SAE with small hidden layers learns denser features. However, this increased feature density leads to poorer performance in distinguishing between target and remaining concepts, as shown in the histogram of reconstruction loss. This may be because larger hidden layers are capable of learning more fine-grained features, which in turn enhance the ability to differentiate between different concepts.

\mbox{}
\clearpage

\section{Additional Qualitative results}

\subsection{Example 1 of celebrities erasure.}
\begin{figure}[!h]
\centering
\includegraphics[width=0.95\textwidth,valign=c]{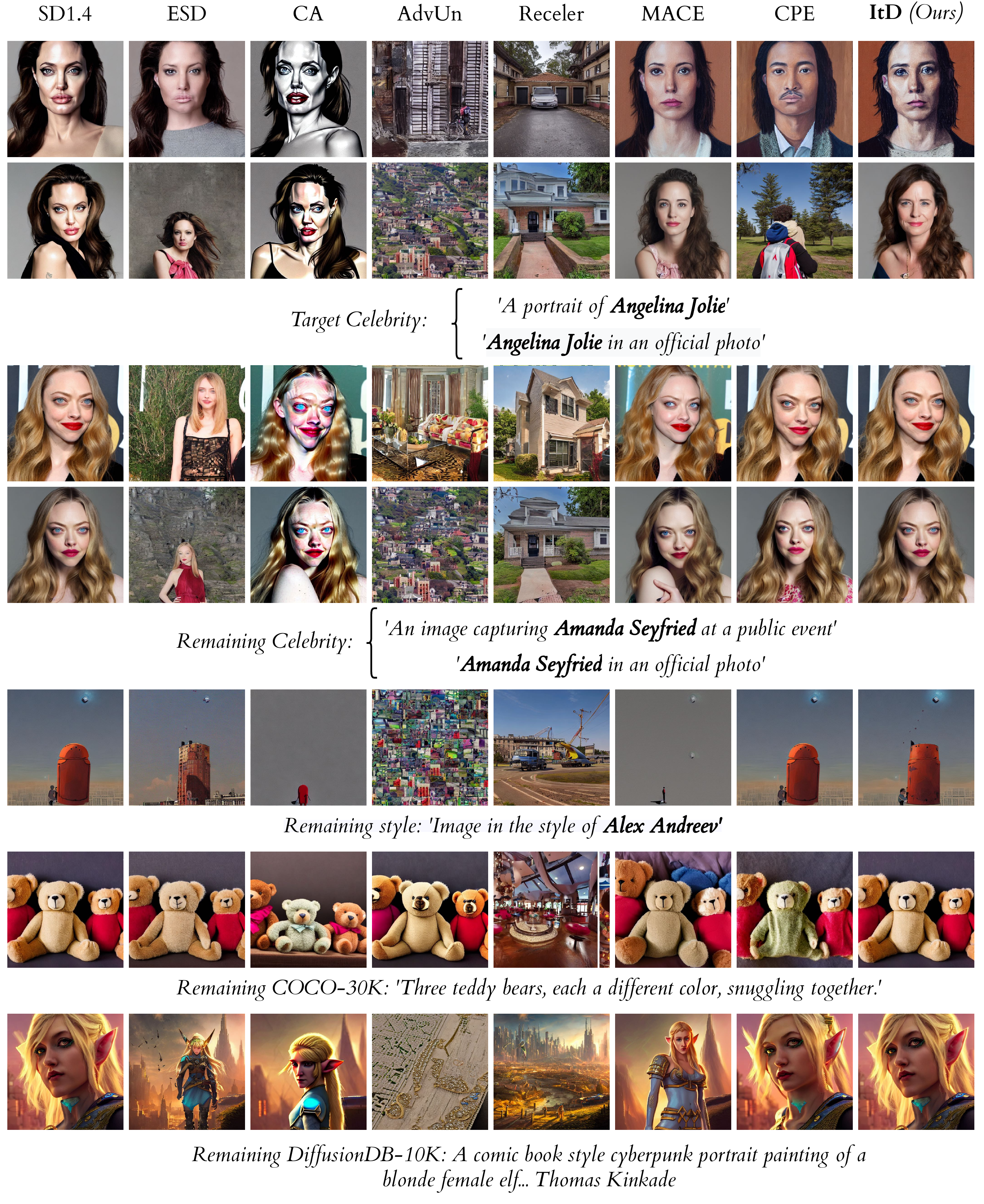}
\caption{Qualitative comparison on celebrities erasure. The images on the same row are generated using the same seed.}
\label{fig:celeb1}
\end{figure}

\subsection{Example 2 of celebrities erasure.}
\begin{figure}[!h]
\centering
\begin{tabular}{c} 
    \subfloat{
        \includegraphics[width=0.97\textwidth,valign=c]{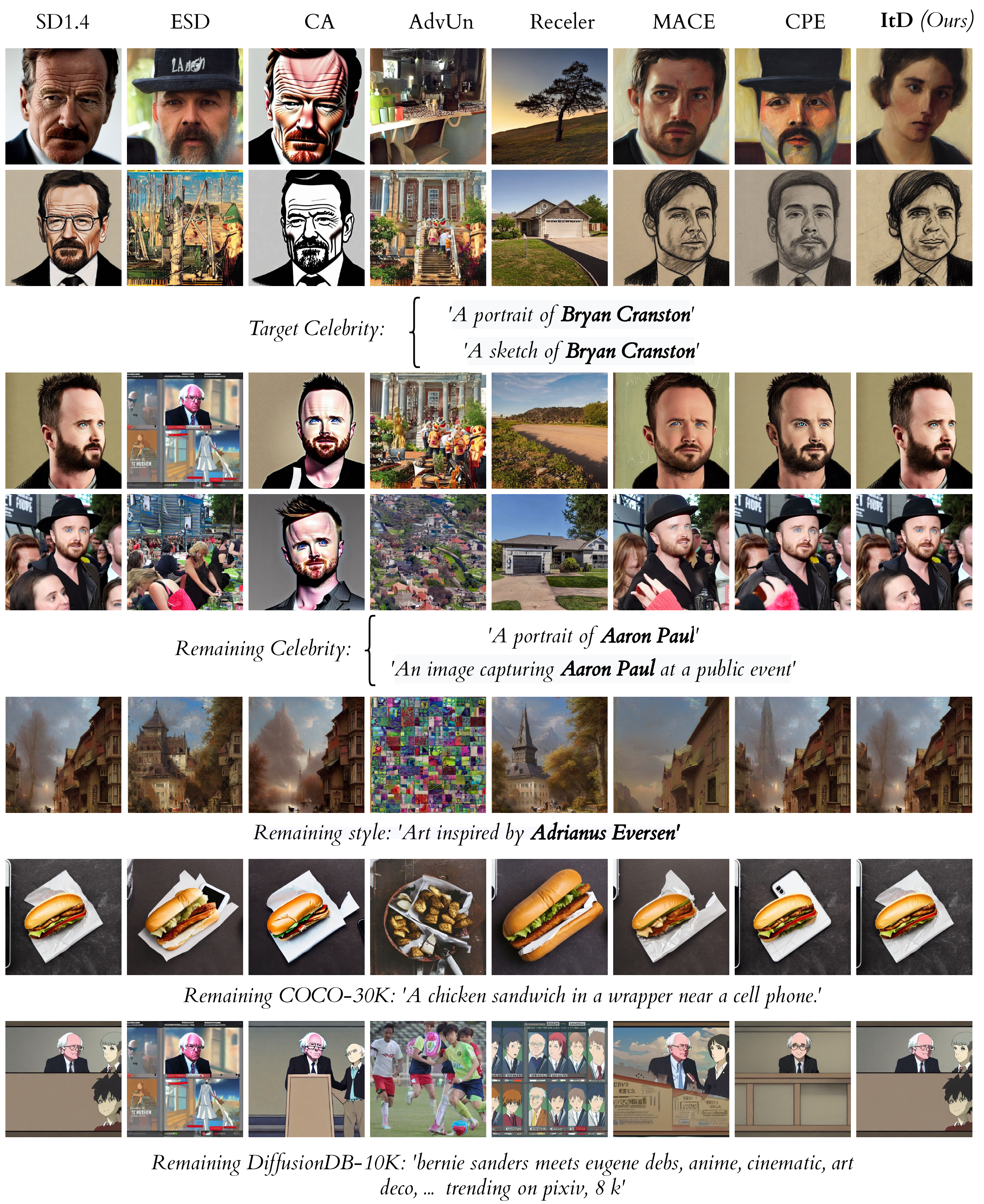}
    } 
\end{tabular}
\caption{Qualitative comparison on celebrities erasure. The images on the same row are generated using the same seed.}
\label{fig:celeb1}
\end{figure}

\subsection{Example 1 of artist styles erasure.}

\begin{figure}[!h]
\centering
\begin{tabular}{c} 
    \subfloat{
        \includegraphics[width=0.98\textwidth,valign=c]{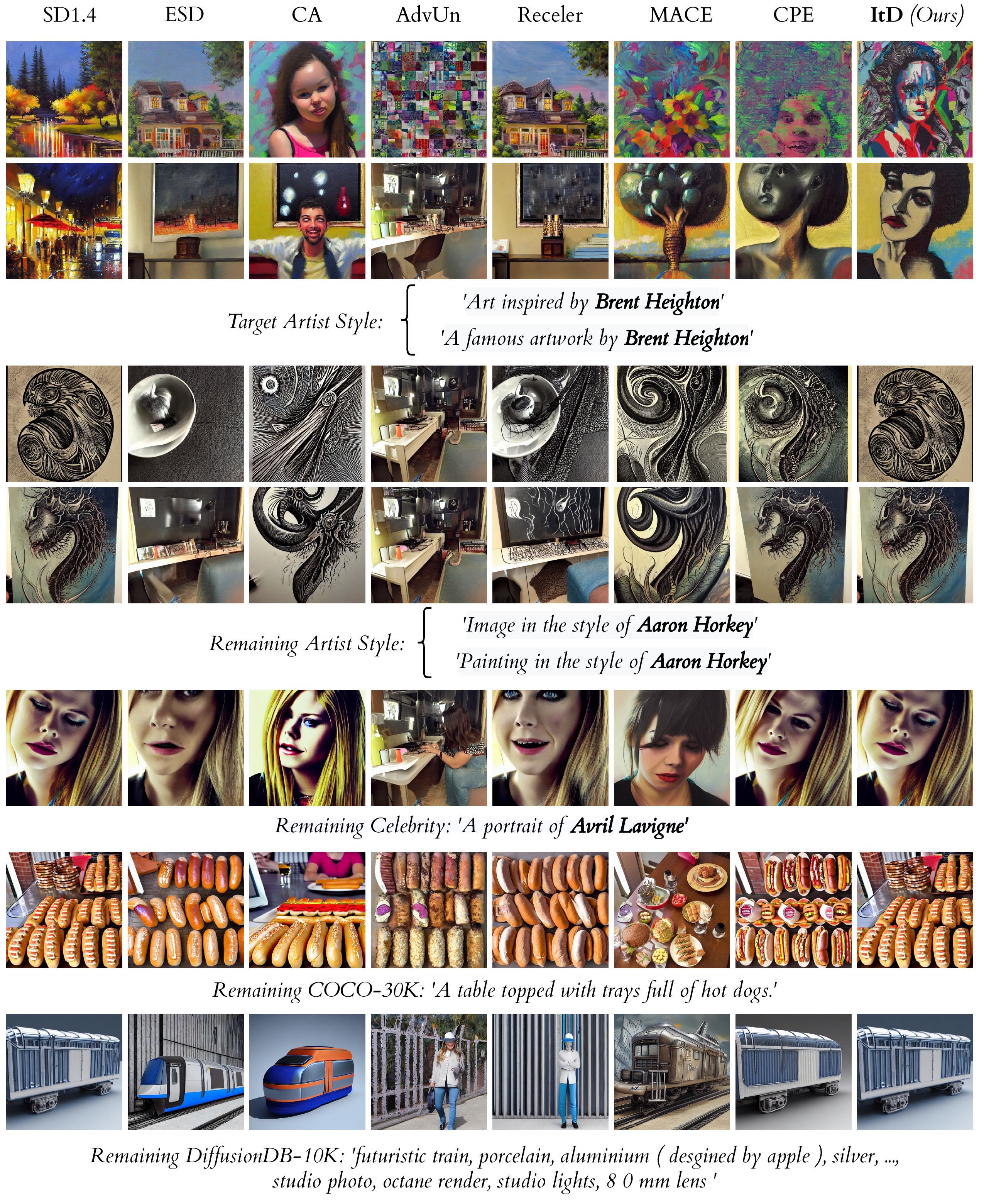}
    } 
\end{tabular}
\caption{Qualitative comparison on artist styles erasure. The images on the same row are generated using the same seed.}
\label{fig:celeb1}
\end{figure}

\subsection{Example 2 of artist styles erasure.}
\begin{figure}[!h]
\centering
\begin{tabular}{c} 
    \subfloat{
        \includegraphics[width=0.96\textwidth,valign=c]{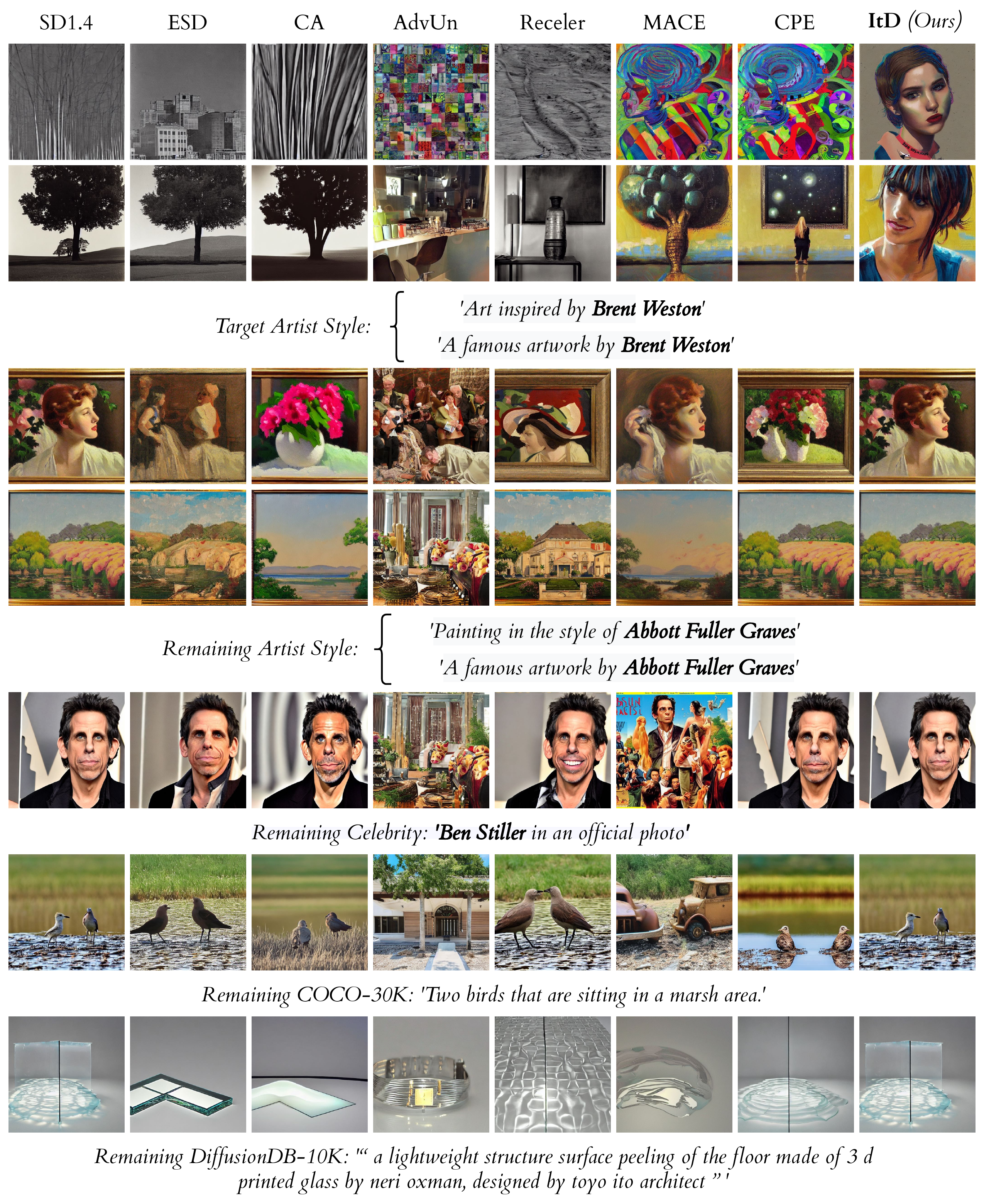}
    } 
\end{tabular}
\caption{Qualitative comparison on artist styles erasure. The images on the same row are generated using the same seed.}
\label{fig:celeb1}
\end{figure}

\end{document}